%% file: 0.main.tex
\documentclass[a4paper,fleqn]{cas-dc}
\usepackage{graphicx}
\usepackage[utf8]{inputenc}
\usepackage{xspace}
\usepackage[table,xcdraw]{xcolor}
\usepackage{booktabs}
\usepackage{multirow} 
\usepackage{amsmath}
\usepackage{amssymb} 
\usepackage{comment}
\usepackage{stmaryrd}
\usepackage{tabularx}

\begin{document}

\newcommand{\etal}{\textit{et al.}\space}
\newcommand{\tool}{$i$\textit{ML}\xspace}

\newcommand{\mlestar}{\textsc{MLE-STAR}\xspace}
\newcommand{\mlzero}{\textsc{MLZero}\xspace}
\newcommand{\automlagent}{\textsc{AutoML-Agent}\xspace}
\newcommand{\autogluon}{\textsc{AutoGluon}\xspace}
\newcommand{\mlebench}{\textsc{MLE-Bench}\xspace}
\newcommand{\imlbench}{\tool-\textsc{Bench}\xspace}

% Short title
\shorttitle{\tool}

% Short author
\shortauthors{ \textit{Le et~al.}}

\title [mode = title]{\tool: Executable, Problem-Grounded, and Broadly Exploratory Code-Driven AutoML}

\author{Dat Le}
[orcid=0009-0002-5745-3361]
\ead{23021519@vnu.edu.vn}
\affiliation{organization={Faculty of Information Technology, VNU University of Engineering and Technology},
    city={Hanoi},
    country={Vietnam}}

\author{Duc-Cuong Le}
[orcid=0009-0007-3748-4810]
\ead{23021485@vnu.edu.vn}

\author{Anh-Son Nguyen}
[orcid=0009-0007-3748-4810]
\ead{23021684@vnu.edu.vn}

\author{Tuan-Dung Bui}
[orcid=0009-0007-7318-6896]
\ead{21020006@vnu.edu.vn}

\author{Thu-Trang Nguyen}
[orcid=0000-0002-3596-2352]
\ead{trang.nguyen@vnu.edu.vn}

\author{Son Nguyen}
[orcid=0000-0002-8970-9870]
\ead{sonnguyen@vnu.edu.vn}\cormark[1]

\author{Hieu Dinh Vo}
[orcid=0000-0002-9407-1971]
\ead{hieuvd@vnu.edu.vn}

% Corresponding author text
\cortext[cor1]{Corresponding author}

\input{1.abstract}

\maketitle

\input{2.intro}
\input{3.background}
\input{4.approach}

\input{5.eval_method}
\input{6.results}
\input{8.conclusion}

\bibliographystyle{elsarticle-num}
\bibliography{references}

\end{document}

%% file: 1.abstract.tex
\begin{abstract}
Automated Machine Learning (AutoML) has improved access to machine learning, yet existing techniques often remain limited in flexibility, transparency, and execution reliability. Code-driven AutoML offers a promising direction by synthesizing executable code for preprocessing, model training, and evaluation. However, current LLM-based approaches frequently generate code that is plausible in text yet brittle in execution, insufficiently grounded in the actual dataset, or restricted to narrow solution paths.
In this paper, we introduce \tool, a multi-agent code-driven AutoML framework designed around three requirements: \textit{executability}, \textit{problem grounding}, and \textit{broad exploration of valid solutions}.
\tool first analyzes the task and profiles the data, then synthesizes a structured blueprint that guides modular code generation across multiple implementation tracks, including traditional ML, pretrained adaptation, and custom neural architectures. To improve reliability, \tool enforces interface checking, dynamic execution, and iterative debugging during integration.
We evaluate \tool on \mlebench and the newly introduced \imlbench, covering diverse Kaggle-style tasks. 
%
% Experimental results show that \tool achieves strong operational reliability and competitive performance across diverse tasks. 
%
On \mlebench, \tool attains a 90\% valid submission rate and a 45\% medal rate, and an APS of 0.82, improving the average standardized performance score (APS) over the LLM-based baselines by 52\%--273\%. On \imlbench, it achieves the highest APS and demonstrates robust performance even when task descriptions are substantially stripped. These results establish \tool as a reliable and competitive framework for code-driven AutoML.

\end{abstract}


%% file: 2.intro.tex
\section{Introduction}

Machine learning (ML) has become a cornerstone of modern technological innovation, driving advancements across diverse domains ranging from healthcare to finance~\cite{ml1,ml2,ml3}. However, the development of effective ML solutions remains a labor-intensive and complex process, typically requiring significant human expertise in data preprocessing, model architecture selection, and hyperparameter optimization~\cite{automl-need1,automl-need2}. As the demand for ML applications outpaces the supply of skilled data scientists, there is a critical need for systems that can autonomously navigate these intricate engineering workflows.

To democratize access to these technologies, Automated Machine Learning (AutoML) has emerged as a pivotal field, aiming to automate the end-to-end ML development pipeline~\cite{auto-sklearn,automl-survey1,automl1,auto-keras,autofe,autogluon,h2o,nas-survey}. By reducing the reliance on manual intervention, AutoML systems seek to accelerate experimentation and enable low/non-experts to deploy robust, data-driven solutions without deep technical prerequisites~\cite{automl-survey1}.
Despite their promise, traditional AutoML systems often face significant limitations regarding flexibility, transparency, and controllability. 
Many existing frameworks are designed as rigid tools tailored for specific tasks, primarily tabular data, limiting their applicability to diverse or multimodal problems. 
Furthermore, these systems typically suffer from a fundamental lack of transparency. Specifically, they operate as ``black boxes'' where the model creation process is obscured, preventing users from observing or understanding how specific architectural decisions were reached. This opacity results in a critical lack of controllability. Consequently, the generated solutions cannot be easily customized or revised, meaning that even minor adjustments to the objective or data often necessitate rebuilding the entire process from scratch.

To address these constraints, recent Large Language Model (LLM)-based agents have shifted towards \textit{Code-driven} AutoML~\cite{mle-star,mlzero,automl-agent}. 
Instead of selecting from a fixed menu of pipelines, a code-driven AutoML method can synthesize executable code for data preprocessing, model training, and evaluation. This shift is attractive because code is inherently more expressive than black-box search over a closed library space. In principle, it enables the system to construct task-specific pipelines, integrate diverse modeling paradigms, and expose the generated solution in a form that users can inspect and revise. However, flexibility alone does not make code-driven AutoML reliable. Existing LLM-based agents often generate solutions that look plausible in text but fail in execution, rely on assumptions about the dataset that are never empirically validated, or remain trapped within narrow and brittle search behaviors such as refining a monolithic script or delegating core reasoning to a fixed AutoML backend.
Search-based systems, such as \mlestar~\cite{mle-star}, rely heavily on retrieving and refining monolithic scripts from existing solutions. This coarse exploration strategy often modifies the entire code structure simultaneously, making it difficult to isolate errors or perform deep exploration within specific components.
Planning-centric systems, such as \automlagent~\cite{automl-agent}, employ a simulation-first approach that relies on text-based pseudo data analysis before code generation. This reliance on textual reasoning without actual execution can lead to failures, where the generated pipeline fails during final implementation due to runtime errors.
Furthermore, \mlzero~\cite{mlzero} operates as \textit{intelligent wrappers} around existing AutoML backbone, \autogluon~\cite{autogluon}. While this ensures executability, it intrinsically limits the system's capabilities to the boundaries of the underlying library, sacrificing the flexibility to architect custom pipelines for complex or non-standard data distributions.

In this paper, we introduce \tool, a novel multi-agent code-driven AutoML framework which is designed to satisfy the three following criteria: (1) \textbf{Executability}: it must reliably synthesize pipelines that run end-to-end and produce valid models or submissions. (2) \textbf{Problem Grounding}: its planning and implementation decisions must be grounded in the full problem context including task-relevant knowledge from the input description, the observed properties of the input data rather than unverified textual assumptions. (3) \textbf{Diverse and Updated Exploration}: it must search across heterogeneous, task-appropriate, and current solution strategies, rather than prematurely committing to a single narrow modeling path.
To ensure executability, \tool verifies intermediate outputs through dynamic execution, interface checking, and iterative debugging. To ensure problem grounding, \tool grounds planning in the task description, empirical profiling of the raw dataset, and retrieved task-relevant knowledge, such as domain constraints and suitable preprocessing strategies, before code synthesis. To ensure broad and up-to-date exploration, \tool combines retrieval-backed planning with modular implementation across multiple modeling tracks, including traditional ML, pretrained adaptation, and custom neural architectures.

Specifically, \tool organizes the workflow into three coordinated phases.  
In \textit{Code-Guided Strategic Planning}, 
% \tool analyzes the task and profiles the dataset to construct an empirical view of the problem. The system then integrates task analysis, empirical data profiling, and retrieved task-relevant knowledge, including domain constraints and preprocessing strategies suited to the data and model, to synthesize a strategic blueprint that specifies preprocessing directives, modeling logic, training protocols, and evaluation requirements. 
%
the system analyzes the task, profiles the dataset, retrieves task-relevant knowledge and model candidates, and synthesizes a strategic blueprint. 
In \textit{Code-Modular Implementation}, specialized coding agents then implement the preprocessing and modeling components separately under explicit interface contracts. 
In \textit{Code-Verifiable Integration}, \tool assembles the generated components, validates them through execution, localizes faults, and iteratively repairs failures until a valid pipeline is obtained. 
%
% This design allows \tool to move beyond both black-box AutoML and monolithic code generation by enforcing reliability while retaining architectural flexibility.

To evaluate the effectiveness of \tool and the existing AutoML approaches, we conducted several experiments on two benchmarks, \mlebench~\cite{mle-bench} and the newly introduced \imlbench.
The experimental results demonstrate that \tool consistently outperforms state-of-the-art code-driven approaches and traditional AutoML baselines. On \mlebench, \tool achieved a valid submission rate of 90\% and a medal rate of 45\%, significantly surpassing baselines like \mlzero (30\% medal rate) and \mlestar (15\% medal rate). Meanwhile, on \imlbench, \tool maintained a dominant overall standardized performance score of 0.54, nearly tripling the performance of \mlzero with standardized performance score of only 0.19 and proving its superior domain adaptation.
Especially, \tool demonstrated remarkable robustness, maintaining a robust 70\% success rate even when task descriptions are significantly stripped.
Collectively, these results establish \tool as a robust paradigm for AutoML, bridging the gap between high-level strategic reasoning and empirically verifiable implementation to achieve reliable, human-competitive performance.

In summary, our main contributions are as follows:
\begin{enumerate}

    \item  \textbf{Novel Design Principles}: We introduce three criteria for ideal code-driven AutoML systems: executability, data grounding, and diverse, updated exploration of valid solutions. These criteria provide a principled lens for analyzing the strengths and limitations of existing approaches.

    \item \textbf{\tool Framework}: We introduce \tool, a multi-agent framework that operationalizes these principles through empirical profiling, retrieval-backed planning, modular code synthesis, dynamic verification, and iterative debugging.
    
    \item \textbf{Extensive Experimental Results}: We demonstrated that \tool consistently outperforms state-of-the-art automated agents on \mlebench and the newly introduced \imlbench, achieving superior operational reliability and competitive medal rates.

\end{enumerate}

The detailed experimental results and source code for reproducing experiments can be found on our website~\cite{website}.

%% file: 3.background.tex
\section{Related Work}

\textbf{Automated Machine Learning (AutoML)}. 
The democratization of machine learning has long been driven by AutoML systems designed to abstract away the complexities of model selection and hyperparameter optimization (HPO)~\cite{automl-survey1,automl1, auto-keras,auto-sklearn,automl-need1,automl-need2}. Early advancements were dominated by optimization-centric frameworks. Auto-WEKA~\cite{auto-weka} and Auto-Sklearn~\cite{auto-sklearn} pioneered the use of Bayesian Optimization to navigate the joint space of data preprocessing and ML algorithms, treating the pipeline configuration as a black-box function to be maximized.
Parallel to general pipeline optimization, specialized subfields have targeted specific bottlenecks. Neural Architecture Search (NAS)~\cite{nas-survey} emerged to automate the architecture design of deep neural networks, modeling the neural architecture construction as a search problem to discover optimal architecture that can outperform human-designed architectures. 
Simultaneously, Automated Feature Engineering~\cite{autofe,openfe} addresses the labor-intensive process of data preparation, employing techniques like expansion-reduction and deep feature synthesis to construct informative predictors from raw data without manual intervention. TPOT~\cite{tpot} bridged these gaps by utilizing genetic programming to evolve tree-based pipeline structures, allowing for more flexible feature engineering combinations.

Recently, the field shifted towards stacking and ensembling strategies to maximize predictive performance. Frameworks like AutoGluon~\cite{autogluon} and H2O AutoML~\cite{h2o} introduced multi-layer stack ensembles and efficient resource allocation strategies, establishing SOTA benchmarks on tabular data. FLAML~\cite{flaml} further optimized this by prioritizing low-cost search strategies to achieve rapid convergence.

Despite their efficacy, these traditional systems operate largely as opaque ``black boxes''. They typically provide limited transparency into how model and pipeline decisions are made, while restricting users to fixed and pre-defined search spaces. This limits interpretability and makes it difficult for domain experts to inspect, customize, or extend the generated solutions beyond standard formats. As a result, these systems are often less suitable for tasks that require flexible engineering decisions or fine-grained manual refinement.

\textbf{LLM-based Multi-Agent for Code-driven AutoML}.
To overcome the rigidity of black-box AutoML, recent studies have shifted toward \textit{Code-driven AutoML}~\cite{automl-gpt,mlcopilot,mle-star,mlzero,automl-agent}, where agents synthesize end-to-end executable pipelines for data preprocessing, model training, and evaluation. This paradigm is promising because code provides a more flexible and inspectable representation of the solution process. Recent systems have explored this direction from several complementary perspectives.

DS-Agent~\cite{ds-agent} employs case-based reasoning to leverage expert solutions from platforms such as Kaggle, enabling low-resource adaptation by reusing successful prior cases. 
Building on the idea of external knowledge acquisition, \mlestar~\cite{mle-star} adopts a Search-as-Tool paradigm, retrieving existing solution scripts from the web and refining them through iterative ablation. This strategy broadens access to candidate solutions, but the retrieved scripts often remain monolithic, making fine-grained debugging and component-level optimization more difficult.

% To mitigate this, \mlestar~\cite{mle-star} adopts a Search-as-Tool paradigm to search on the web instead of the case bank. Rather than synthesizing architectures from scratch, they retrieve existing solution scripts from the web and refine them via iterative ablation studies. While effective for standard tasks, this approach suffers from \textit{monolithic rigidity}. Refining a single retrieved script typically involves coarse exploration strategies that modify the entire code structure simultaneously. This entanglement makes it difficult to isolate errors or perform deep, targeted optimization of specific components without destabilizing the rest of the pipeline.
Frameworks such as \automlagent~\cite{automl-agent} focus on retrieval-augmented planning to decompose tasks into manageable subproblems. This improves high-level coordination, but the planning process may still rely heavily on pseudo-analysis through textual reasoning rather than execution over the actual data, which can weaken the connection between the generated plan and runtime behavior.

Recent systems such as \mlzero~\cite{mlzero} incorporate cognitive perception and memory modules to support more complex and multimodal ML workflows. However, these systems often rely on strong existing backbones, such as \autogluon~\cite{autogluon}, for core model construction and execution. This provides strong executability, but also ties the explored solution space to the capabilities of the underlying library.

In contrast, \tool is designed around three criteria for ideal code-driven AutoML: \textit{executability}, \textit{data grounding}, and \textit{diverse and updated exploration of valid solutions}. \tool grounds planning in empirical profiling of the input data, explores multiple implementation tracks (e.g., traditional ML, pretrained adaptation, and custom neural architectures) and enforces executability through interface checking, dynamic execution,  and iterative debugging. In this way, \tool combines the flexibility of code-driven synthesis with stronger runtime reliability and broader solution exploration.

% addresses these limitations through a unified architectural response. By employing \textit{Code-Guided} blueprinting, we eliminate the strategic ambiguity found in retrieved scripts. Furthermore, by enforcing a \textit{Code-Verifiable} workflow that executes code at every intermediate stage, we directly mitigate the risks of hallucinated logic inherent in simulation-first methodologies. Through a \textit{Code-Modular} implementation, we resolve the monolithic rigidity that prevents fine-grained debugging, and by synthesizing native architectures, we transcend the performance ceilings of library-dependent wrapper approaches.

%% file: 4.approach.tex
\section{Executable, Data-Grounded, and Broadly Exploratory Code-Driven AutoML}

\input{4.approach/0.problem_principles}

\input{4.approach/1.executablity}
\input{4.approach/2.data_grounding}
\input{4.approach/3.sol_space}

\input{4.approach/4.integration}

%% file: 4.approach/0.problem_principles.tex
\subsection{Problem Formulation and Design Principles}
In \tool, we formalize AutoML as a constrained code generation task. Given a task description $\mathcal{T}$ and an available training dataset $\mathcal{D}_{train}$, the objective is to synthesize an optimal executable pipeline $\mathcal{S}^*$ that maximizes a performance metric $\mathcal{M}_\mathcal{T}$ (e.g., accuracy, F1-score) on held-out or evaluation data $\mathcal{D}_{test}$, subject to logical and resource constraints $\mathcal{C}$. Formally:
%
% \begin{equation}
%     \mathcal{S}^* = \operatorname*{argmax}_{\mathcal{S} \in \mathbb{S}} \mathcal{M}_\mathcal{T}(\llbracket \mathcal{S} \rrbracket(\mathcal{D})) \quad \text{s.t.} \quad \mathcal{V}_{exec}(\mathcal{S}) = \text{True}
% \end{equation}
% %
\begin{equation}
    \mathcal{S}^* =
    \operatorname*{argmax}_{\mathcal{S} \in \mathbb{S}}
    \mathcal{M}_{\mathcal{T}}
    \left(
        \llbracket \mathcal{S} \rrbracket(\mathcal{D}_{train}, \mathcal{D}_{test})
    \right)
\end{equation}
such that $ \mathcal{V}_{exec}(\mathcal{S}; \mathcal{D}_{train}, \mathcal{D}_{test}) = \text{True}$, where $\mathbb{S}$ is the space of syntactically valid scripts, and
$\llbracket \mathcal{S} \rrbracket$ denotes the execution of script
$\mathcal{S}$, including training on $\mathcal{D}_{train}$ and producing
predictions or submissions for $\mathcal{D}_{test}$. Meanwhile,
$\mathcal{V}_{exec}(\cdot)$ denotes a runtime verification function that
checks execution integrity, data-flow consistency, and validity of the
generated outputs. Note that final generalization performance is measured only on $\mathcal{D}_{test}$.
Unlike a purely optimization-centric view, however, we argue that a high-performing code-driven AutoML system must satisfy three requirements simultaneously: \textit{it must generate pipelines that are executable, it must ground its decisions in the full problem context with input data and domain knowledge, and it must explore a diverse and updated space of valid solutions rather than committing prematurely to a narrow modeling path}.

\begin{figure*}
    \centering
    \includegraphics[width=1.0\linewidth]{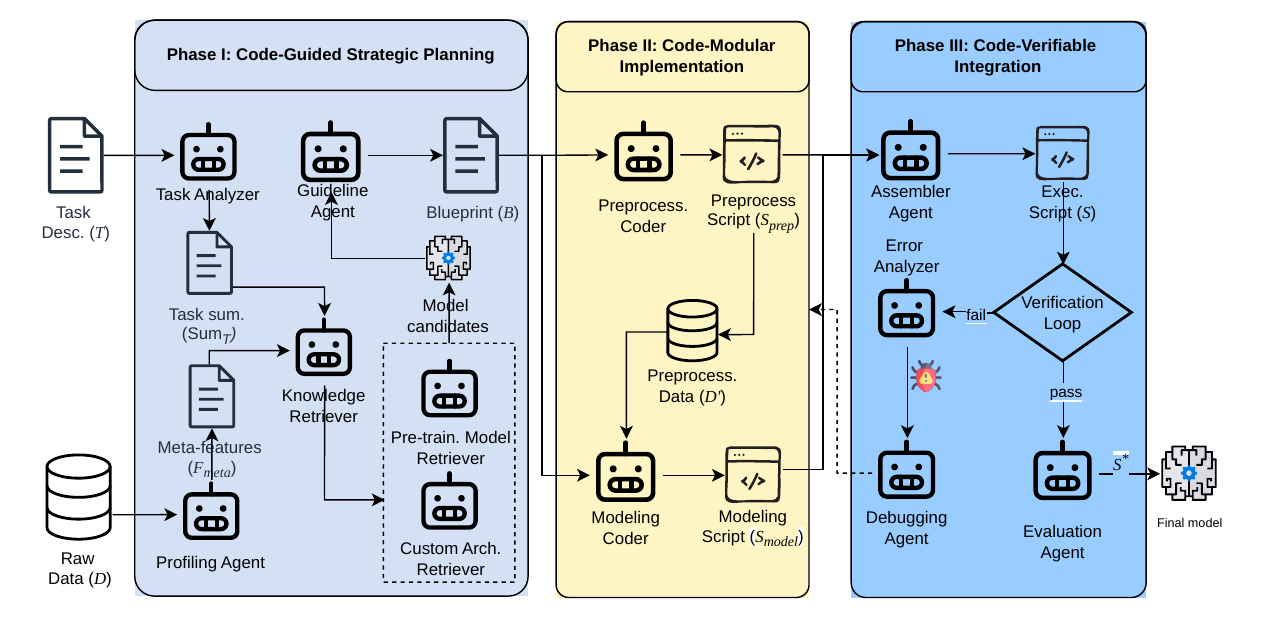}
    \caption{Architectural overview of \tool}
    \label{fig:framework_overview}
\end{figure*}

\tool is designed explicitly around these requirements. 
Figure~\ref{fig:framework_overview} provides an overview of the framework, organized into three phases: \textit{Code-Guided Strategic Planning}, \textit{Code-Modular Implementation}, and \textit{Code-Verifiable Integration}. 
Rather than relying on direct-to-code prompting or generating a monolithic script in a single pass, \tool decomposes the AutoML workflow into these coordinated phases.
In Phase I, \textit{Code-Guided Strategic Planning}, \textit{Task Analyzer} first analyzes the task semantics and \textit{Profiling Agent} empirically profiles the raw dataset, and dedicated retrievers acquire up-to-date model candidates and task-relevant knowledge, including domain constraints and preprocessing strategies.  
Based on these inputs, \textit{Guideline Agent} then synthesizes a strategic blueprint that specifies preprocessing directives, model-selection logic, training protocols, and evaluation requirements. 
In Phase II, \textit{Code-Modular Implementation}, specialized coding agents implement preprocessing and modeling components under explicit interface contracts.
This modularization avoids entangled direct-to-code generation and enables each component to be developed, checked, and repaired independently.
In Phase III, \textit{Code-Verifiable Integration}, an assembly and debugging loop integrates the components, validates them through execution, localizes errors, and iteratively repairs failures until a valid executable pipeline is obtained. 
In this way, \tool enforces executability, problem grounding, and broad solution exploration through phase-level coordination rather than isolated agent behavior. 
The remainder of this section explains the architecture from this perspective.

%% file: 4.approach/1.executablity.tex
\subsection{Principle I: Executability}

The first and most fundamental requirement of code-driven AutoML is \textbf{\textit{executability}}. A generated code is not useful if it is only textually plausible but fails during preprocessing, training, or evaluation. Accordingly, \tool treats executability not as a post hoc check, but as a first-class design constraint enforced throughout generation and integration.

\subsubsection{Interface and Execution Contracts as Executable Constraints}
To enforce executable compatibility between independently generated components, \tool adopts a contract-driven architecture. 
These contracts are not limited to the boundary between preprocessing and modeling. Rather, they define execution-oriented constraints across agents and phases, including whether coding agents follow the strategic blueprint, whether intermediate data artifacts satisfy expected schemas, whether generated components can be integrated consistently, and whether the final output satisfies task-specific requirements such as producing a valid \texttt{submission.csv} with the required columns.

A central contract in \tool is the interface contract $\text{IC}_\mathcal{T}$ between the preprocessing and modeling components.
During planning, Guideline Agent formulates an interface contract $\text{IC}_\mathcal{T}$ which specifies the exact object types and output structures expected between preprocessing and modeling stages, such as a \texttt{pandas.DataFrame}, a \texttt{torch.DataLoader}, or a configuration object. This contract functions as a binding constraint on downstream generation and is used to prevent type mismatches, incompatible tensor shapes, invalid feature schemas, or other cross-module inconsistencies before they propagate into unrecoverable failures.

At the preprocessing stage, the contract acts as a rigid postcondition $Q _{\text{post}}$: the generated preprocessing script is required that the dataset $\mathcal{D'}$, transformed from the raw dataset $\mathcal{D}$, satisfies the exact schema constraints required by the downstream modeling stage, including shape, type, and validity properties such as non-nullness or correctly encoded categorical values. At the modeling stage, the same contract is reinterpreted as a precondition $R_{\text{pre}}$, ensuring that the generated architecture is mathematically compatible with the verified structure of $\mathcal{D'}$. 

% By treating the preprocessing output as a hard type and schema constraint, \tool decouples the modeling decision from fragile LLM-based static inference over code and instead grounds it in a deterministic interface specification.

Beyond this component-level interface, \tool also enforces phase-level execution contracts. For example, the generated code must implement the steps specified in the blueprint, the integrated script must preserve data-flow consistency across modules, and the final execution must produce valid task artifacts such as trained models, predictions, or submission files. By treating these requirements as explicit executable constraints rather than implicit assumptions in LLM-generated code, \tool reduces fragile static inference and grounds implementation in verifiable specifications.

\subsubsection{Dynamic Integration and Intermediate Execution Verification}

Executability in \tool is enforced through dynamic integration rather than simple textual concatenation. 
In \tool, the target pipeline $\mathcal{S}$ is decomposed into two main generated components: a preprocessing component $\mathcal{S}_{\text{prep}}$, which transforms the raw dataset into model-ready representations, and a modeling component $\mathcal{S}_{\text{model}}$, which defines training, validation, inference, and output generation logic.
Although these components are generated separately for modularity, they must be assembled into a single executable pipeline to preserve end-to-end data flow, support reproducible execution, and produce a deployable inference script without relying on large intermediate files stored on disk.

In the final integration phase, Assembler Agent acts as the system integration engine. It combines the preprocessing script $\mathcal{S}_{\text{prep}}$ and the modeling script $\mathcal{S}_{\text{model}}$ into a unified executable pipeline $\mathcal{S}$. 
Particularly, Assembler Agent performs semantic variable binding between the output variables of preprocessing and the input variables of modeling, while also resolving dependencies, imports, and serialization logic. This preserves data-flow continuity across independently synthesized code blocks and improves reproducibility of the resulting pipeline.

Crucially, \tool enforces intermediate execution verification before finalizing the pipeline. The framework executes $\mathcal{S}_{\text{prep}}$ on the given dataset $\mathcal{D}$ to obtain the processed representation $\mathcal{D'}$, then validates the runtime properties of $\mathcal{D'}$, including shapes, data types, and sparsity patterns, against the interface contract $\text{IC}_\mathcal{T}$. 
The integrated pipeline $\mathcal{S}$ is further executed on the input data to test physical feasibility before full training or inference proceeds. 
This mechanism empirically checks whether the observable postconditions of preprocessing ($Q_{\text{post}}$) are compatible with the modeling preconditions ($R_{\text{pre}}$), rather than assuming compatibility from the generated code alone. While execution cannot prove all semantic properties of the pipeline, it substantially reduces the risk that syntactically plausible but incompatible code progresses to later stages.

% that the implication $Q_{\text{post}} \implies R_{\text{pre}}$ is validated empirically rather than assumed. This prevents semantically invalid pipelines from progressing simply because they appear syntactically correct.

\subsubsection{Fault Localization and Iterative Debugging}

If execution fails, \tool invokes an Error Analyzer and a Debugging Agent. 
Rather than relying on generic retry behavior, Error Analyzer uses the runtime traceback together with the architectural invariants, interface contract, and execution context to localize the failure.
%
%
% Errors are categorized into two classes: contract fulfillment failure, where preprocessing fails to produce an output satisfying the required postcondition, and contract usage violation, where the modeling logic violates the input assumptions defined by the contract.
%
The failures handled by \tool are not limited to interface violations; they may include syntax errors, missing dependencies, incompatible library APIs, incorrect file paths, invalid preprocessing logic, tensor-shape mismatches, resource constraints, or invalid prediction and submission formats. Contract-related failures are treated as an important subclass, including contract fulfillment failures, where an upstream component fails to produce the expected postcondition, and contract usage violations, where a downstream component violates the assumptions specified by the contract.

Once the fault is localized, Debugging Agent repairs the relevant component iteratively until the execution constraint is satisfied. For the failures caused by outdated APIs, library changes, or incomplete internal knowledge, Debugging Agent is equipped with a web search capability for retrieving current syntax and documentation. This targeted repair mechanism avoids full regeneration when possible and limits regression to the smallest necessary scope. As a result, \tool enforces executability through a closed-loop process of assembly, execution, diagnosis, and correction until $\mathcal{V}_\text{exec}(\mathcal{S})=\text{True}$.

%% file: 4.approach/2.data_grounding.tex
\subsection{Principle II: Problem Grounding}

Executability alone is insufficient for credible code-driven AutoML. The generated code may run successfully and still be semantically wrong if its logic is built on hallucinated assumptions about the task or the data. This issue becomes particularly severe when task descriptions are ambiguous, incomplete, or under-specified, as is common in realistic automation settings. For this reason, \tool treats \textit{\textbf{problem grounding}} as a second core requirement: the system must anchor its planning and implementation decisions in the full problem context, including task description with the retrieved relevant knowledge and the observed properties of the dataset rather than relying solely on textual reasoning.

\subsubsection{Task Analysis and Empirical Profiling, and Knowledge Retrieval}

The Task Analyzer first constructs a task summary $Sum_\mathcal{T}$ from the raw task description $\mathcal{T}$, extracting semantic constraints such as the optimization metric, expected submission format, and any explicit problem instructions. 
However, \tool does not assume that this textual description is complete or fully reliable. To prevent the propagation of hallucinated data properties, a Profiling Agent executes both a verified data profiling tool and LLM-generated profiling code on the raw dataset $\mathcal{D}$, producing a set of empirical meta-features $F_\text{meta}$. These include observed schema information, feature types, missingness patterns, data quality properties, and sample distributions.
In parallel, a Knowledge Retriever acquires task-relevant knowledge, including domain constraints and preprocessing strategies suited to the characteristics of the data and the target model.

This design is essential in under-specified or noisy settings. By grounding the pipeline in task semantics, observed data properties, and retrieved task-relevant knowledge, \tool reduces the risk that downstream agents will hallucinate nonexistent columns, incompatible feature types, or unsupported preprocessing assumptions. More generally, it allows the system to infer part of the problem structure directly from the dataset when the user-provided task description is ambiguous or incomplete.

\subsubsection{Problem-Grounded Blueprint Synthesis}

The task summary $Sum_\mathcal{T}$, empirical meta-features $F_\text{meta}$, and retrieved task-relevant knowledge $K_\mathcal{T}$, extracted by Task Analyzer, Profiling Agent, and Knowledge Retriever are passed to a Guideline Agent, which synthesizes a strategic blueprint: 
$$B = \langle B_\text{prep}, B_\text{model}, B_\text{train}, B_\text{eval}  \rangle$$
Specifically, the blueprint $B$ defines: 

\begin{itemize}
    \item A preprocessing strategy $B_\text{prep}$: including transformations such as imputation, encoding, scaling, and feature preparation, strictly tailored to the data types and observed properties identified in $F_\text{meta}$ and informed by retrieved preprocessing knowledge in $K_\mathcal{T}$.

    \item A model selection logic $B_\text{model}$: defining the model-selection logic and selects among implementation tracks such as traditional ML, pretrained adaptation, or custom neural architectures, based on the task semantics in $Sum_\mathcal{T}$, the observed data characteristics in $F_\text{meta}$, and relevant modeling knowledge in $K_\mathcal{T}$.

    \item Training protocols $B_\text{train}$: specifying optimization protocols, including the hyperparameter search space, optimization strategy, and loss-function requirements.

    \item An evaluation protocol $B_\text{eval}$: defining validation strategies and performance metrics aligned with the semantics extracted from the task and domain-specific evaluation constraints retrieved in $K_\mathcal{T}$.
    
\end{itemize}

The blueprint could be considered as a structured, problem-grounded set of architectural invariants. By fixing these invariants prior to code generation, \tool prunes the search space from the full script space $\mathbb{S}$ to a constrained subspace $\mathbb{S}_{\mathcal{B}} \subset \mathbb{S}$, where candidate scripts are required to remain compatible with observed data properties, retrieved knowledge, and explicit interface requirements. This yields two benefits: \textit{(i)} it reduces the probability of invalid or semantically inconsistent generation, and \textit{(ii)} it guides the coding agents toward task-appropriate strategies that are justified by the actual structure of the input data and relevant task-relevant knowledge.

\subsubsection{Contract-Aware Preprocessing as Grounded Implementation}

The Preprocessing Coder implements data grounding by transforming the raw dataset $\mathcal{D}$ into a verified processed representation $\mathcal{D'}$ according to the directives in $B_\text{prep}$. This agent implements data loading, cleaning, feature engineering, and format conversion under the postconditions imposed by the interface contract. Importantly, the correctness of the preprocessing stage is not inferred abstractly; the generated script $\mathcal{S}_\text{prep}$ is executed immediately so that the resulting artifacts can be inspected and checked against the contract.
This design ensures that problem grounding is preserved not only at the planning level but also at the level of concrete implementation. The resulting preprocessing module is empirically validated against the actual dataset on which the model will ultimately be trained. 

%% file: 4.approach/3.sol_space.tex
\subsection{Principle III: Broad Exploration of Valid Solutions}

A reliable code-driven AutoML system should not only produce valid pipelines; it should also avoid premature commitment to a narrow solution family. Because ML tasks span heterogeneous modalities and inductive biases, \tool is designed to explore a \textbf{\textit{broad}} and \textit{\textbf{up-to-date space}} of valid solutions before selecting the best executable candidate.

\subsubsection{Multi-Track Candidate Pipeline Generation}

Guided by the blueprint ($B_\text{model}$), \tool explores three complementary implementation tracks in parallel: \textit{traditional ML}, \textit{pretrained model adaptation}, and \textit{custom neural architectures}. Traditional ML supports robust solutions for structured tabular settings; pretrained adaptation leverages transfer learning for standard high-dimensional modalities such as vision and NLP; and custom neural architectures provide flexibility for specialized or non-standard distributions.

The Modeling Coder synthesizes candidate pipelines under these tracks according to the selected architectural directives and training protocols ($B_\text{train}$). 
When appropriate, generated training scripts include embedded hyperparameter optimization routines such as cross-validation loops, Bayesian optimization, or grid search, as dictated by the blueprint $B$. This ensures that diversity is not limited to model family selection alone, but extends to optimization strategy and training configuration. 
By explicitly allowing multiple implementation tracks, \tool avoids the dependency ceiling of wrapper-based systems and the rigidity of approaches that commit too early to a single modeling path.

\subsubsection{Retrieval-backed Expansion of Solution Space}

To support broad and up-to-date exploration, \tool uses retrieval to expand the candidate solution space beyond the backbone LLM's internal knowledge. Instead of relying only on models or implementation patterns already known to the LLM, \tool retrieves external resources that can guide different modeling tracks. These retrieved resources fall into three categories.

\textit{First}, for pretrained-model tracks, the \textit{Pretrained Model Retriever} searches for compatible checkpoints from external repositories such as Hugging Face. These checkpoints provide reusable model candidates for tasks where transfer learning is appropriate.
\textit{Second}, for custom deep-learning tracks, the \textit{Custom Architecture Retriever} searches for architecture patterns suited to the empirical properties of the dataset identified during profiling. For example, it may retrieve CNN-style architectures for visual data or Transformer-style architectures for sequence-based tasks.
\textit{Third}, \textit{Knowledge Retriever} collects task-relevant knowledge, including preprocessing strategies, domain constraints, evaluation conventions, and implementation patterns suited to the data characteristics and target model.

This retrieval-backed design serves two purposes. First, it broadens exploration beyond a fixed AutoML library or a single model family. Additionally, it helps keep the explored solution space current by incorporating recent architectures, checkpoints, and implementation guidance. As a result, \tool can generate candidate pipelines that are both more diverse and better aligned with the task context.

% \tool includes a Resource Retrieval Layer. For pretrained tracks, a Pretrained Model Retriever uses web search to identify compatible checkpoints from external repositories such as Hugging Face. 
% %
% For custom deep learning tracks, a Custom Architecture Retriever searches for model architectures suited to the empirical properties of the dataset identified during profiling, such as CNN families for visual data or Transformer-style architectures for sequence-based tasks. 
% %
% In addition, a Knowledge Retriever acquires task-relevant knowledge, including preprocessing strategies, domain constraints, and implementation patterns suited to the characteristics of the data and the target model.
% %
% This retrieval-backed design serves two purposes. First, it broadens the solution space beyond a fixed internal library or a single model family. Second, it keeps the search process updated by allowing the framework to incorporate current architectures, checkpoints, and implementation guidance rather than relying exclusively on static internal knowledge of the backbone LLMs.

\subsubsection{Evaluation, Selection, and Optional Ensembling}

Once candidate pipelines have passed executable verification, \tool compares them empirically. The Evaluation Agent computes the task-specific performance metric $\mathcal{M}_\mathcal{T}$ on the validation protocol specified by $B_\text{eval}$, and the system selects the globally best validated solution $\mathcal{S}^*$ for final deployment. 
In addition, \tool also supports modular ensembling. Because the Assembler Agent treats models as interchangeable modules that satisfy a common interface contract, the system can synthesize an ensemble pipeline $\mathcal{S}_\text{ens}$ that aggregates the predictions of the top-$k$ validated models through mechanisms such as voting or stacking. This provides a further level of exploratory flexibility without requiring manual reconfiguration of the end-to-end workflow. While the framework defaults to selecting the strongest validated candidate, the architecture is explicitly designed to support broader aggregation when beneficial.

%% file: 4.approach/4.integration.tex
\subsection{Integrated Workflow}
The three requirements in \tool are enforced jointly. Problem grounding constrains planning to strategies consistent with the actual dataset. Broad exploration expands the candidate space across multiple valid and current solution families. Executability then filters this space through assembly, runtime verification, and iterative debugging until a valid end-to-end pipeline is obtained. The result is a system that does not merely generate code, but synthesizes, validates, and selects among executable ML solutions under empirical and architectural constraints.

Operationally, \tool can still be viewed as proceeding through three coordinated stages: planning, modular implementation, and verifiable integration. However, these stages are not independent design goals. They are the mechanism by which the framework realizes the more fundamental requirements of executable synthesis, problem-grounded reasoning, and diverse, updated exploration.

%% file: 5.eval_method.tex
\section{Evaluation Methodology}
To evaluate the effectiveness of \tool, we seek to answer the following research questions (RQs):

\textbf{RQ1. Performance Analysis:} How effective is \tool in building ML models compared to the AutoML baselines?
    
\textbf{RQ2. Intrinsic Analysis:} How do different components of \tool contribute to its overall performance?
    
\textbf{RQ3. Sensitivity Analysis:} How sensitive is \tool's performance to different external factors?

\textbf{RQ4. Efficiency Analysis:} How efficient is \tool in terms of execution time and computational cost?

\input{5.eval_method/benchmarks}
\input{5.eval_method/procedure}

\input{5.eval_method/metrics}

%% file: 5.eval_method/benchmarks.tex
\subsection{Benchmarks}
To evaluate the efficacy of \tool, we conducted experiments across two distinct benchmarks designed to test generalization and robustness.

\textbf{\mlebench (Lite)}~\cite{mle-bench}. 
We first evaluated its performance on 20 datasets from the Lite subset of \mlebench, a computationally feasible proxy for the full benchmark. Although not identical, \mlebench reports a high correlation in system rankings between the Lite and full versions, making Lite a practical basis for comparison.

\textbf{\imlbench}.
To further evaluate the methods, we introduced \imlbench, which is a curated benchmark consisting of 16 recently launched Kaggle competitions (2021--2025), distinct from those in \mlebench and covering diverse domains and problem types.
The competitions were selected based on three rigorous standards to ensure quality and relevance:

\begin{enumerate}
    \item \textit{Recency and Popularity}: Most selected competitions received substantial community participation, and we prioritized competitions with at least 1,000 submissions when available.

    \item \textit{Diversity}: The benchmark spans multiple domains and modalities, including tabular learning, natural language, computer vision, audio, object detection, and multimodal prediction.

    \item \textit{Quality Assurance}: We manually verified the accessibility of all data files and the determinism of the evaluation metrics.
\end{enumerate}

To prevent data leakage and simulate a realistic deployment scenario, we split each original training set into benchmark training and test subsets with an 80/20 ratio. All studied methods were restricted to the training subset for model development, and evaluation is performed strictly on the held-out test subset.
The details of \imlbench could be found on the project's website~\cite{website}.

%
% The overall statistics of 

%% file: 5.eval_method/procedure.tex
\subsection{Evaluation Procedure}

\textbf{RQ1. Performance Comparison}. We compared \tool against the traditional \autogluon baseline and the three state-of-the-art code-driven agents: 
\begin{itemize}
    \item \autogluon~\cite{autogluon}: A state-of-the-art traditional (non-LLM) AutoML library. We include this baseline to determine whether code-driven agents can surpass the performance ceiling of standard ``black-box'' AutoML tools, while also comparing against a system that is strong in executability but limited in architectural flexibility and breadth of exploration.
    
    \item \mlzero~\cite{mlzero}: A multi-agent framework utilizing cognitive perception and dual-memory modules. Notably, \mlzero employs \autogluon as its core execution engine, effectively functioning as an LLM-driven wrapper for an existing library rather than generating native model architectures. This provides a useful comparison point for assessing whether broader solution exploration beyond a fixed backend yields measurable gains.
    
    \item \mlestar~\cite{mle-star}: A search-based agent that leverages a ``Search-as-Tool'' paradigm. It retrieves existing solution scripts from the web to form an initial solution and subsequently refines them through iterative ablation studies targeting specific code blocks. This baseline is particularly relevant for comparing against \tool's exploration capability, while also highlighting the limitations of monolithic script refinement.
    
    \item \automlagent~\cite{automl-agent}: A planning-centric framework that employs retrieval-augmented planning. It decomposes the pipeline into sub-tasks and relies on prompting-based simulation (pseudo-analysis) to guide code generation before implementation. This baseline is especially relevant for evaluating the value of grounding planning in empirical data rather than textual simulation alone.

\end{itemize}

To ensure fair comparison, we followed the similar evaluation procedure as in the existing studies~\cite{mle-star}. Particularly, all LLM-based agents utilized the same backbone model, Gemini-2.5-Flash. Experiments were conducted on a standardized Kaggle environment equipped with a single NVIDIA P100 GPU.
We enforced a strict resource constraint of 5 hours per dataset for the end-to-end process (code generation, training, and inference). For \mlzero and \automlagent, this constraint was enforced via system prompts and timeout interruptions. For \mlestar, we set the hyperparameters to 2 model candidates and 2 refinement rounds to fit within the time budget.
To ensure statistical reliability, each method was executed across 3 independent runs with distinct random seeds, and the mean for all metrics is reported.
For the non-LLM \autogluon, we used a standardized implementation protocol based on the official API and documentation, with fixed resource limits matching the same runtime budget. 
%
% For each task, \autogluon was given the same training files, target specification, and time budget as the agentic systems; no task-specific manual feature engineering beyond the standardized protocol was used.
%
We report this baseline as a strong library-based reference point.

\textbf{RQ2. Intrinsic Analysis}. To quantify the contribution of our three core design principles, we evaluated \tool against three degraded architectural variants. This isolates the impact of Executability, Data Grounding, and Diverse and Updated Exploration of Valid Solutions

\textbf{\textit{(A) Executability-oriented analysis}}. 
We first analyze the mechanisms that enable \tool to produce runnable pipelines and valid models under realistic runtime constraints.

\begin{enumerate}
     \item \textit{\tool w/o verification ($\tool_{\text{static}}$):} The intermediate execution and runtime verification steps are disabled. The system relies solely on syntactic checks by LLMs, mimicking simulation-first approaches. $\tool_{\text{static}}$ tests the necessity of dynamic verification for ensuring runtime feasibility and valid pipeline execution.
    
    \item \textit{\tool w/ mono code ($\tool_{\text{mono}}$):} The \textit{Preprocessing} and \textit{Modeling} agents are merged into a single one that generates a monolithic script, removing Assembler's role in dynamic integration. This variant evaluates the role of modular decomposition in improving executable integration, error isolation, and repairability.

    \item \textit{Impact of Self-Correction Budget:} \textit{Debug Agent} iteratively repairs code upon verification failure. We varied the maximum allowed debugging rounds $K \in \{0, 1, 3, 5, 10, 20\}$ to analyze the trade-off between \tool's executability, performance, and token/time cost.
\end{enumerate}

\textit{\textbf{(B) Problem-grounding-oriented analysis. }}
We next analyze whether \tool's strategic decisions benefit from being explicitly grounded in the full problem context, including the task description, observed data properties, and retrieved task-relevant knowledge.

\begin{itemize}

    \item \textit{Impact of Data Profiling:} We compared \tool with $\tool_{\text{reactive}}$, a \tool variant without data profiling. In this setting, the system relies only on the task description without empirical analysis of the dataset. This analysis evaluates whether explicit data profiling improves the system's ability to construct task-appropriate pipelines and avoid unsupported assumptions about the input data.

    \item \textit{Impact of Knowledge Retrieval:} We further evaluated a degraded variant without the Knowledge Retriever to analyze the contribution of retrieved task-relevant knowledge. In this setting, the system relies only on the task description and empirical profiling results without external retrieval of preprocessing strategies, domain constraints, or implementation guidance. This analysis evaluates whether retrieved knowledge improves the system's ability to construct task-appropriate pipelines and avoid narrow or unsupported implementation choices.

\end{itemize}

\textit{\textbf{(C) Exploration-oriented analysis.}}
Finally, we analyze the mechanisms that enable \tool to search over a broad and up-to-date space of valid solutions instead of committing early to a single modeling path.

\begin{itemize}
    \item \textit{Impact of Multi-track Exploration:} The blueprint allows for a \textit{hybrid} search space (Traditional ML, Custom NN, Pretrained). We compared this against restrictive configurations where the agent is forced to use only \textit{single-implementation-track} strategies. This analysis evaluates whether broader exploration across heterogeneous model families improves performance.

    \item \textit{Impact of Search Engine / Retrieval Backend:} Since \tool's exploration capability partly depends on external retrieval, we evaluated how the retrieval configuration affects its ability to discover diverse and up-to-date valid solutions. In particular, this analysis examines the contribution of external search to identifying relevant architectures, pretrained checkpoints, and implementation patterns beyond the internal prior knowledge of the backbone LLM.

    \item \textit{Impact of Aggregation Logic:} We compared the default \textit{single-best selection} strategy against a \textit{model ensemble} strategy (modular ensembling is enabled). This analysis evaluates whether aggregating multiple validated candidates is more effective than selecting the single strongest explored pipeline.

    \item \textit{Impact of Backbone LLM:} We also replaced Gemini-2.5 with LLMs of varying capabilities, including \textit{GPT-4o mini}, \textit{GPT-5 mini}, and \textit{Qwen 2.5 Coder 7b}. This analysis tests how the backbone model affects \tool's ability to reason over, implement, and refine candidate solutions drawn from a broad search space.
\end{itemize}

\textbf{RQ3. Sensitivity Analysis}. We assessed the robustness of \tool by varying input factors to determine if our architectural benefits persist beyond the default configuration:

\begin{itemize}

    \item \textit{Task Description Granularity:} We evaluated the system's resilience to ambiguous user intent. For a subset of tasks, we stripped the detailed Kaggle description down to a sparse prompt (e.g., \textit{``Analyze the provided data files and execute the appropriate predictive task for the target variable''}). 
    % This tests whether \textit{Code-Guided Planning} can correctly infer objectives and metrics via empirical profiling alone, effectively filling the information gap.

    \item \textit{ML Data Modality}: We investigated the performance of \tool by distinct data modalities including \textit{Natural Language}, \textit{Vision}, \textit{Tabular}, \textit{Audio}, and \textit{Multimodal}.
\end{itemize}

%% file: 5.eval_method/metrics.tex
\subsection{Metrics}
For \textit{\mlebench}, we report the six standard metrics defined by the MLE-Bench protocol:
\begin{itemize}
    \item \textbf{\%Valid:} The ratio of successfully executed and valid submitted runs over the total number of runs.
    
    \item \textbf{\%Above Median:} The percentage of submissions whose scores exceed the median score of all participants on the corresponding private leaderboard of the Kaggle competition.
    
    \item \textbf{\%Medal (Gold/Silver/Bronze):} The fraction of submissions that achieve results comparable to respective medal Kaggle leaderboard positions.
    
    \item \textbf{\%Any Medal:} The combined proportion of submissions that fall into any of the medal-winning categories (gold, silver, or bronze).
\end{itemize}

For both \mlebench and \imlbench, 
Because the benchmark includes heterogeneous competition metrics (e.g., LogLoss, F1, RMSE, Kappa), we map raw scores to a unified normalized performance score, $P_{\text{norm}} \in [0, 1]$, to facilitate aggregation. Let $Sc$ be the raw score achieved by an agent. The normalized performance score is calculated as follows:

\begin{enumerate}
    \item \textbf{Higher-is-Better Metrics} ${\in [0, 1]}$: For metrics such as AUC, F1-score, Accuracy: $P_{\text{norm}} = Sc$
    
    \item \textbf{Lower-is-Better Metrics}: For unbounded loss metrics (e.g., MAE, LogLoss), we apply exponential decay transformation to map the error to $[0, 1]$:
    $P_{\text{norm}} = e^{-Sc}$.
    Here, a raw loss of $0$ yields a score of $1.0$.

    \item \textbf{Bounded Metrics} $\in [-1, 1]$: For metrics such as Quadratic Weighted Kappa or Matthews Correlation Coefficient:
    $P_{\text{norm}} = \frac{Sc + 1}{2}$

    \item \textbf{Execution Failure}: If an agent fails to produce a valid submission (e.g., runtime error, timeout) or produces invalid ones, the score is penalized as: $P_{\text{norm}} = 0$
    
\end{enumerate}
To evaluate the overall performance of a method across the entire benchmark, the final \textit{Average Performance Score}, $\textit{APS}$, is calculated by taking the average of the normalized scores across all $N$ competitions:
$$\textit{APS} = \frac{1}{N} \sum_{i=1}^{N} P^i_{\text{norm}}$$
We use APS only for aggregate comparison and retain task-level scores to avoid hiding metric-specific behavior.

%% file: 6.results.tex
\section{Experimental Results}

\input{6.results/RQ1}
\input{6.results/RQ2}
\input{6.results/RQ3}
\input{6.results/RQ4}
\input{6.results/threats}

%% file: 6.results/RQ1.tex
\subsection{RQ1. Performance Comparison}

Tables~\ref{tab:rq1-comparsion-mlebench} and~\ref{tab:rq1-comparsion-iml} show the performance of the SOTA code-driven AutoML agents, as well as traditional \autogluon baseline on \mlebench and \imlbench, respectively. 
Overall, \textit{\tool consistently achieved superior results in terms of operational reliability and competitive performance across two benchmarks with diverse task types}.

\input{tables/rq1-mlebench}

On the \mlebench, \tool demonstrated a significant improvement in operational robustness, achieving a valid submission rate of 90\%. 
The equality between made and valid submissions indicates that all submissions produced by \tool were valid.
This underscores the efficacy of \textit{Code-Verifiable} integration, which employs dynamic execution to identify and corrects runtime errors before finalization. Meanwhile, \mlestar and \automlagent exhibited lower valid submission rates of 55\% and 40\% respectively, indicating that monolithic refinement strategies and simulation-first approaches frequently suffer from unrecoverable runtime hallucinations.

Furthermore, \tool showed a competitive advantage, with 80\% of its submissions exceeding the human median score on Kaggle. This suggests that \tool's \textit{Code-Guided} planning does not merely aim for executability but effectively prunes the search space toward high-quality architectural decisions. This advantage was achieved by systematically exploring three complementary implementation tracks and grounding the neural/pretrained tracks with retrieved state-of-the-art architectures or pretrained models, thereby pruning the search space toward high-quality designs. Our approach secured any medal (Gold, Silver, or Bronze) in 45\% of the tasks, outperforming the state-of-the-art traditional AutoML tool, \autogluon, which achieved a 40\% medal rate. This achievement is particularly notable as it demonstrates that \tool can transcend the performance ceiling of ``black-box'' optimization by architecting native, task-specific pipelines rather than relying on fixed library wrappers (\mlzero).

\input{tables/rq1-imlbench}

The results on \imlbench further validate \tool's domain adaptation and robustness across non-standard data distributions. As shown in Table \ref{tab:rq1-comparsion-iml}, \tool achieved an overall average normalized performance score (APS) of 0.54, significantly higher than its closest competitors, \mlestar (0.36) and \autogluon (0.39). While traditional baselines like \autogluon performed strongly on tabular tasks like \textit{Steel Plate} (0.89) or \textit{Paddy Disease} (0.96), they did not support or struggle significantly with complex or multi-label tasks. For instance, in \textit{PetFinder} and \textit{Multilabel Classification} tasks, most baselines failed to produce meaningful results (scoring 0.00), whereas \tool maintained high scores of 0.67 and 0.69, respectively.

This performance gap is consistent with the benefits of \tool's core design principles of problem grounding, executability, and broad exploration of valid solutions. Particularly, \tool grounds its strategic blueprint in empirical meta-feature profiling and retrieved task-relevant knowledge, then implements the solution through separate \textit{Preprocessing Coder} and \textit{Modeling Coder} modules.
This design supports both sophisticated feature engineering and custom model construction beyond the predefined search space of traditional wrappers (e.g., \mlzero). In contrast, \mlzero, which relies on \autogluon as its backbone, collapsed to 0.00 on 9 of 16 tasks, illustrating the dependency ceiling of wrapper-based methods. By combining grounded planning with modular implementation and execution-aware integration, \tool is better positioned to discover and realize task-appropriate algorithms for diverse settings.

Overall, these results indicate that the strength of \tool lies in the interaction of its three core design principles, including problem grounding, executability, and broad exploration of valid solutions, which together support a more reliable and flexible code-driven AutoML framework.

\textbf{Result Analysis}. To better understand the strength and limitations of the approaches, we further analyzed the ML tasks in both benchmarks.

A failed case for all evaluated methods, including \tool, occurred during \textit{MLSP 2013 Bird Classification} task from \mlebench. The goal is to identify bird species in 10-second audio clips within noisy environments (e.g., wind, rain, or buzzing insects).
This multi-label audio classification task is challenging because methods have to extract sophisticated features from noisy acoustic environments and correctly synchronize disparate data artifacts, including raw waveforms, metadata, and label mappings.
However, the task description provided insufficient clarity regarding the specific ID mapping between audio files and labels. 
Since even \textit{Guideline Agent} relies on textual descriptions and initial profiling produced by \textit{Task Analyzer} and \textit{Profiling Agent} to infer join logic, it synthesized a flawed strategic blueprint. This flaw was propagated to \textit{Preprocessing Coder}, which subsequently generated defective join logic that resulted in unrecoverable runtime exceptions.
Thus, this case underscores a fundamental limitation of LLM-based AutoML: regardless of modular isolation or code-verifiable safeguards, the system's operational integrity highly relies on the clarity and completeness of the initial problem semantics.

In contrast, for \textit{PetFinder (Adoption Pred.)}\footnote{PetFinder - Adoption Prediction: \textit{https://www.kaggle.com/compe titions/petfinder-adoption-prediction}} from \imlbench, \tool achieved a score of 0.67 while all the others failed ($P_{norm} = 0$).
This task requires the integration of diverse data modalities, including tabular metadata, image directories, and unstructured text descriptions, linked through a common \texttt{PetID}.
The other methods frequently failed in such settings because the entanglement of data loading and feature engineering creates complex bugs that exceed the recovery capabilities of standard automated debugging.

\tool overcame these limitations by combining problem-grounded planning, modular implementation, and execution-aware verification..
Table~\ref{tab:blueprint_alignment} shows the simplified blueprint generated in \textit{Code-guided Planning} and corresponding implementation generated in \textit{Code-Modular}.
Unlike the agents attempting to process all data at once, \tool's blueprint explicitly decouples the storage of ID (\texttt{PetID}) for submission at Step 2 from the separation of the target variable (\texttt{AdoptionSpeed}) at Step 3. This ensures that the implementation code is physically isolated from the modeling logic, preventing data leakage and reducing cognitive load on the coding agents.
In this task, \tool identified that raw text descriptions are incompatible with the chosen tabular neural model. Our method proactively created \texttt{description\_length} and \texttt{num\_words\_description} (Step 5). By specifying the exact transformation logic in the guideline, the resulting code (e.g., \texttt{X\_train['Description'].apply(len)}) is designed to produce features aligned with the downstream model requirements.
The blueprint enforces a rigid batch-processing contract (Step 9), requiring a \texttt{batch\_size} of 64 and the use of \texttt{DataLoader} objects. If the loader fails, Debugging Agent can fix the specific containing function (\texttt{preprocess\_data}) without needing to rewrite the architecture, a level of fault isolation that monolithic agents cannot achieve.

Overall, this example illustrates how \tool uses grounded planning, modular code generation, and execution-aware verification to bridge raw multimodal data and an executable modeling pipeline. This allowed \tool to achieve an APS of 0.67, while other agents failed to bridge the gap between raw data and executable modeling pipelines.

\input{tables/example}

\textit{\textbf{Interestingly}}, we observed that \tool discovered non-obvious, high-performing strategies that diverge from standard human and baseline conventions, in \textit{Leaf Classification} task from \mlebench.
This task is originally categorized as image classification, as the dataset includes images of leaf specimens alongside pre-extracted tabular features.

Many developers often build CNN-based models to process raw image data\footnote{Here is an example: \textit{https://www.kaggle.com/code/quangnguynv/leaf-classification-cnn/notebook}}. 
Similarly, \mlzero attempted an image-centric approach by defaulting to the multi-modal backend of \autogluon. Although intuitive, these standard paths often yield suboptimal results due to the high noise in the raw image data and the superior information density of the pre-extracted features. 
Meanwhile, \mlestar selected a LGBM-based approach focused on standard tabular processing. \automlagent correctly identified the tabular path, yet implemented a shallow, under-optimized MLP (only two hidden layers).

\tool's \textit{Guideline Agent} identified a more effective (yet unconventional) strategy by prioritizing the extracted feature set over raw image processing. 
By executing three parallel strategic tracks, \tool discovered that a custom Deep Multi-Layer Perceptron (MLP) anchored on the tabular data significantly outperformed both standard computer vision models and generic tabular libraries.
\tool synthesized a Deep MLP involving specialized components:

\begin{itemize}
    \item A sequential structure with \texttt{BatchNorm}, \texttt{ReLU}, and \texttt{Dropout} layers to optimize for distinct species classes.

    \item The calculation of inverse frequency class weights to prevent bias toward dominant species.

    \item A stratified data split by strict \texttt{StandardScaler} fitting only on the training split to prevent label leakage .
\end{itemize}

By bypassing the obvious computer vision route in favor of a deeply engineered tabular neural pipeline, \tool demonstrated architectural creativity that exceeded both human-authored baselines and simpler automated agents. 
This case illustrates how \tool can use data-grounded planning and multi-track exploration to select a less obvious but empirically stronger modeling path.

%% file: tables/rq1-mlebench.tex
\begin{table*}
\centering
\caption{Performance Comparison on \mlebench}
\label{tab:rq1-comparsion-mlebench}

\begin{tabular}{l|rrr|rrrr|r}\toprule
\textbf{} &Made Sub. &Valid Sub. &Above Median &Bronze &Silver &Gold &Any Medal &$\textit{APS}$ \\\midrule
\autogluon &80\% &80\% &55\% &0\% &\textbf{10\%} &30\% &40\% &0.66 \\\cmidrule{1-9}
\mlzero &70\% &60\% &45\% &\textbf{5\%} &5\% &20\% &30\% &0.54 \\\cmidrule{1-9}
\mlestar &65\% &55\% &40\% &0\% &0\% &15\% &15\% &0.43 \\\cmidrule{1-9}
\automlagent &50\% &40\% &25\% &0\% &0\% &5\% &5\% &0.22 \\\cmidrule{1-9}
\tool &\textbf{90\%} &\textbf{90\%} &\textbf{80\%} &0\% &\textbf{10\%} &\textbf{35\%} &\textbf{45\%} &\textbf{0.82} \\
\bottomrule
\end{tabular}
\end{table*}

%% file: tables/rq1-imlbench.tex
\begin{table*}
\centering
\caption{Performance Comparison on \imlbench}
\label{tab:rq1-comparsion-iml}
\begin{tabular}{lrrrrr}
\toprule
\textbf{Competition} & \autogluon & \mlzero & \mlestar & \automlagent & \tool \\
\midrule
Predict Eff. Arg. & 0.46 & 0.44 & 0.45 & 0.44 & \textbf{0.47} \\
Domain Classify & 0.44 & 0.44 & 0.41 & \textbf{0.45} & 0.40 \\
Predict LLM & 0.25 & 0.22 & 0.10 & 0.00 & \textbf{0.26} \\
PetFinder (Adoption Pred.) & 0.00 & 0.00 & 0.00 & 0.00 & \textbf{0.67} \\
Multilabel Cls. & 0.00 & 0.00 & 0.57 & 0.00 & \textbf{0.69} \\
Plant Traits & 0.00 & 0.00 & 0.00 & 0.00 & \textbf{0.01} \\
Dog Breed & 0.85 & 0.00 & 0.45 & 0.49 & \textbf{0.90} \\
Paddy Disease & 0.96 & 0.00 & 0.94 & \textbf{0.97} & \textbf{0.97} \\
Steel Plate & \textbf{0.89} & \textbf{0.89} & \textbf{0.89} & \textbf{0.89} & \textbf{0.89} \\
Synthetic object detection & 0.00 & 0.00 & 0.00 & 0.00 & \textbf{0.90} \\
CZII CIRO & \textbf{0.04} & 0.00 & 0.00 & 0.00 & 0.00 \\
Tensorflow Speech Recognition & 0.93 & 0.00 & 0.62 & 0.00 & \textbf{0.98} \\
Lmsys & \textbf{0.36} & 0.35 & 0.35 & 0.00 & 0.33 \\
MABe & 0.00 & 0.00 & 0.00 & 0.00 & \textbf{0.14} \\
Store Sales & \textbf{0.65} & 0.47 & 0.60 & 0.00 & 0.62 \\
Child Mind Institute & 0.39 & 0.30 & 0.34 & 0.00 & \textbf{0.41} \\
\midrule
\textit{\textbf{Overall APS}} & 0.39 & 0.19 & 0.36 & 0.20 & \textbf{0.54} \\
\textit{\textbf{Valid Submission}} & 0.69 & 0.44 & 0.69 & 0.31 & \textbf{0.94} \\
\bottomrule
\end{tabular}
\end{table*}

% https://docs.google.com/spreadsheets/d/11rWpSZJdHk56ulwIZYiWF95gP7_pRkcl58DscqdDDQc/edit?gid=764780035#gid=764780035

% \begin{table*}
% \caption{Performance comparison on iML-Bench.}
% \begin{tabular}{lccccc}
% \toprule
% \textbf{Dataset}  & \textbf{iML (Ours)} & \textbf{MLZero} & \textbf{MLE Star} & \textbf{AutoML Agent} \\
% \midrule
% Predict Eff. Arg.  & \textbf{0.47} & 0.44 & 0.45 & 0.44 \\
% Domain Classify  & 0.4 & 0.44 & 0.41 & \textbf{0.45} \\
% Predict LLM  & \textbf{0.26} & 0.22 & 0.1  & 0.0 \\
% PetFinder  & \textbf{0.67} & 0 & 0  & 0 \\
% Multilabel Cls.  & \textbf{0.69} & 0 & 0.57 & 0 \\
% Plant Traits  & \textbf{0.01} & 0 & 0 & 0 \\
% Dog Breed  & \textbf{0.9} & 0 & 0.45  & 0.49 \\
% Paddy Disease  & \textbf{0.97} & 0 & 0.94 & \textbf{0.97} \\
% Steel Plate  & 0.89 & 0.89 & 0.89 & 0.89 \\
% Overall Avg.  & \textbf{0.58} & 0.22 & 0.42 & 0.35 \\
% \bottomrule
% \end{tabular}%
% % }
% \end{table*}

%% file: tables/example.tex
\begin{table*}
\centering
\caption{Strategic Blueprint and Implementation Code (simplified) by \tool for \textit{PetFinder}}
\label{tab:blueprint_alignment}
\begin{tabularx}{\textwidth}{@{} c >{\raggedright\arraybackslash}X >{\ttfamily\arraybackslash}X @{}}
\toprule
\textbf{\#Step} & \textbf{Strategic Blueprint} & \textbf{Corresponding Implementation (simplified)} \\ \midrule

1 
& Load training and test datasets into pandas DataFrames. 
& train\_df = pd.read\_csv(paths\_dict["train.csv"]) \newline test\_df = pd.read\_csv(paths\_dict["test.csv"]) \\ \addlinespace
%-------------------------------------
2 
& Store 'PetID' for submission; drop from feature sets. 
& train\_df = train\_df.drop('PetID', axis=1) \newline test\_df = test\_df.drop('PetID', axis=1) \\ \addlinespace
%-------------------------------------
3 
& Separate 'AdoptionSpeed' as the target variable (y). 
& X\_train = train\_df.drop('AdoptionSpeed', axis=1) \newline y\_train = train\_df['AdoptionSpeed'] \\ \addlinespace
%-------------------------------------
4 
& Impute missing 'Name' with the string 'Unknown'. 
& X\_train['Name'] = X\_train['Name'].fillna('Unknown') \newline X\_test['Name'] = X\_test['Name'].fillna('Unknown') \\ \addlinespace
%-------------------------------------
5 
& Create 'descr\_length' and 'num\_words\_desc'; drop raw text. 
& X\_train['desc\_length'] = X\_train['Description'] .apply(len) \newline 
X\_train['num\_words\_desc'] = X\_train['Description']. apply(lambda x: len(x.split())) \newline 
X\_train = X\_train.drop('Description', axis=1) \newline 
//Do the same for X\_test \\ \addlinespace
%-------------------------------------
6 
& Apply LabelEncoder to categorical features for embedding creation. 
& le = LabelEncoder() \newline le.fit(combined\_cat\_data[col].unique()) \newline X\_train[col] = le.transform(X\_train[col]) \\ \addlinespace
%-------------------------------------
7 
& Apply StandardScaler to all identified numerical features. 
& scaler = StandardScaler() \newline X\_train[num\_cols] = scaler.fit\_transform(X\_train[num\_cols]) \\ \addlinespace
%-------------------------------------
8 
& Split data into training and validation sets (8:2 with random\_state=42). 
& X\_train\_split, X\_val, y\_train\_split, y\_val = train\_test \_split(..., test\_size=0.2, random\_state=42) \\ \addlinespace
%-------------------------------------
9 
& Prepare PyTorch DataLoaders with a batch\_size of 64. 
& train\_loader = DataLoader(train\_dataset, batch\_size=64, shuffle=True) \\ \bottomrule
\end{tabularx}
\end{table*}

%% file: 6.results/RQ2.tex
\subsection{RQ2. Intrinsic Analysis}

\subsubsection{Executability-oriented Analysis}

We first analyze the components that help \tool generate pipelines that are physically feasible, executable, and recoverable under runtime failures.

\begin{figure}
    \centering
    \includegraphics[width=1.0\linewidth]{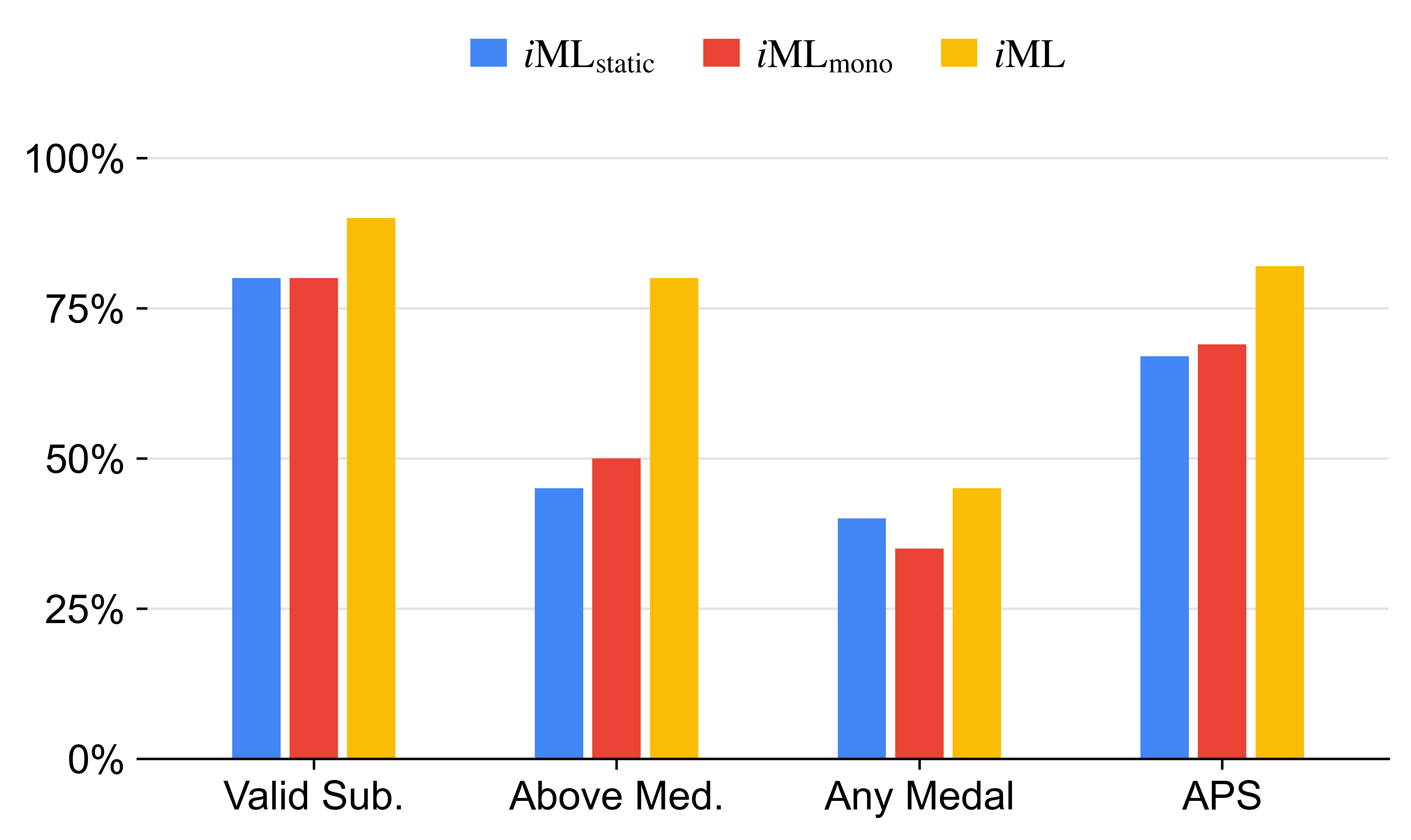}
    \caption{Performance of \tool's degraded variants on \mlebench.}
    \label{fig:iml_variants}
\end{figure}

\textbf{Effect of Runtime Verification and Modular Integration.}
To quantify the contribution of these mechanisms, we evaluated \tool against degraded architectural variants, including \tool$_{\text{static}}$ (\textit{No verification}) and \tool$_{\text{mono}}$ (\textit{Mono code}). Figure~\ref{fig:iml_variants} shows the performance of these variants on \mlebench, illustrating the critical impact of executability-oriented design choices on the framework's overall efficacy.

The \tool$_{\text{mono}}$ (\textit{Mono code}) variant, which merges preprocessing and modeling processes into a \textit{single} monolithic script, shows the risks of logic entanglement. This variant observed a reduction in both the valid submission rate (80\%) and APS (0.69). In a monolithic structure, error propagation becomes a significant bottleneck. For example, a minor syntax or logical error in the early data-loading phase can destabilize the entire pipeline. By contrast, \tool's modular design enforces component isolation through specialized agents, allowing \textit{Assembler Agent} to perform precise variable binding and enabling \textit{Debugging Agent} to localize faults within specific scopes. The drop in \textit{Above Median} rate to 50\% in the monolithic variant further shows that the lack of modular isolation hinders the iterative refinement of individual components, leading to less robust global solutions.

Disabling runtime verification in the \tool$_{\text{static}}$ (\textit{No verification}) variant resulted in the lowest APS (0.67) and a reduced success rate of 80\%. This variant mimics simulation-first approaches that rely on the LLM's internal syntactic checks rather than empirical execution. Our results show a critical hallucination gap: \textit{without intermediate execution to check interface contract on data samples, the coding agents often propagate semantically invalid logic, such as referencing non-existent columns or incompatible tensor shapes}. Full \tool's ability to achieve an 80\% \textit{Above Median} compared to only 45\% for the static variant shows that actual code execution is essential for anchoring the generated pipeline in the reality of the data. By enforcing dynamic verification at key integration points, \tool reduces the propagation of hallucinated or incompatible logic, ensuring that the final integrated pipeline is not only syntactically correct but empirically feasible.

Overall, these results indicate that executability in \tool is not supported by a single component, but by the interaction of modular decomposition, explicit integration, and dynamic verification. Removing either modularity or execution checking leads to substantial degradation in both operational reliability and competitive quality.

\textbf{Impact of Self-Correction Budget.}
In this experiment, we investigated the dynamic relationship between autonomous repair cycles and overall performance of \tool by varying the maximum allowed debugging rounds $K \in \{0, 1, 3, 5, 10, 20\}$.

\begin{figure}
    \centering
    \includegraphics[width=1.0\linewidth]{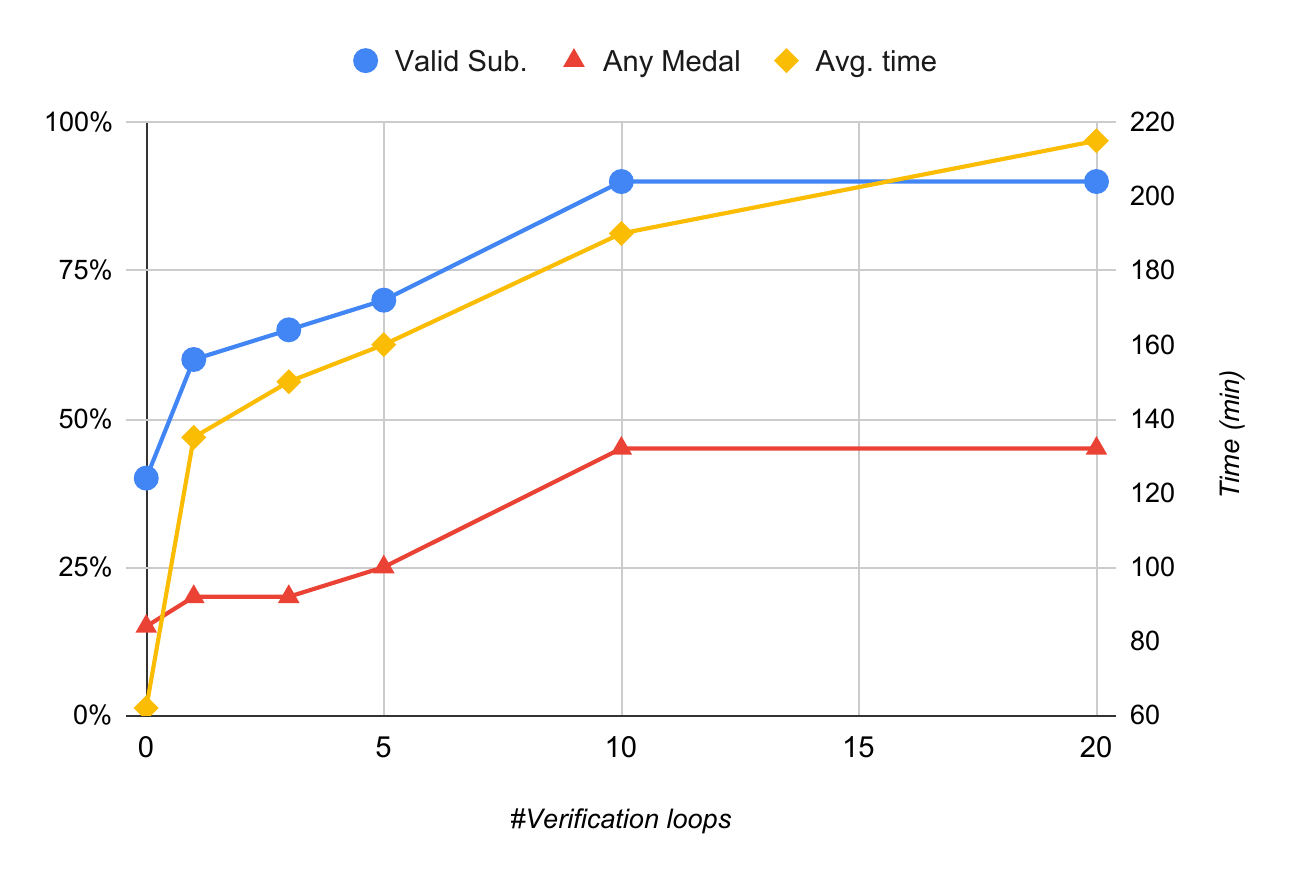}
    \caption{Performance of \tool with different self-correction budgets on \mlebench.}
    \label{fig:debug_rounds}
\end{figure}

As shown in Figure~\ref{fig:debug_rounds}, at $K=0$ (zero-shot implementation without feedback), \tool achieved a valid submission rate of only 40\% and an \textit{Any Medal} rate of 15\%. This variant represents the substantial probability of runtime failure when LLMs are tasked with complex data science workflows without the safeguard of empirical verification. However, even a single round of debugging ($K=1$) resulted in a 50\% relative increase in successful executions (60\% valid submissions) and a 80\% increase in \textit{Above Median} performance (45\% vs. 25\%). This \textit{shallow} repair phase typically addresses routine syntactic errors or deprecated API calls, anchored by the execution-verification component that identifies physical feasibility early in the implementation phase.

As the budget increases to $K=10$, we observed the most significant gains in competitive quality. The \textit{Above Median} rate climbs sharply to 80\%, accompanied by a peak \textit{Any Medal} rate of 45\%. This suggests that higher debugging budgets enable the agent to resolve deeper structural inconsistencies, such as tensor shape mismatches or logic errors within \textit{Preprocessing Coder}'s data transformation logic. Notably, the performance plateaus between $K=10$ and $K=20$, where the valid submission rate and \textit{Any Medal} remain static while average execution time increases 13\%.

The results suggest that setting $K=10$ is the most efficient configuration for \tool, balancing the high computational cost of iterative LLM calls against the diminishing returns of late-stage refinement. This self-correction mechanism helps ensure that the final integrated pipeline satisfies the strategic blueprint's contract, preventing potentially catastrophic runtime failures.

\subsubsection{Problem-grounding-oriented Analysis}

\textbf{Impact of Data Profiling.}
To quantify the contribution of data grounding, we evaluated \tool$_{\text{reactive}}$ (\textit{No data profiling}), a degraded variant in which the \textit{Guideline Agent} no longer has access to empirical data profiling and must rely primarily on the raw task description when guiding downstream code generation.

\begin{figure}
    \centering
    \includegraphics[width=1.0\linewidth]{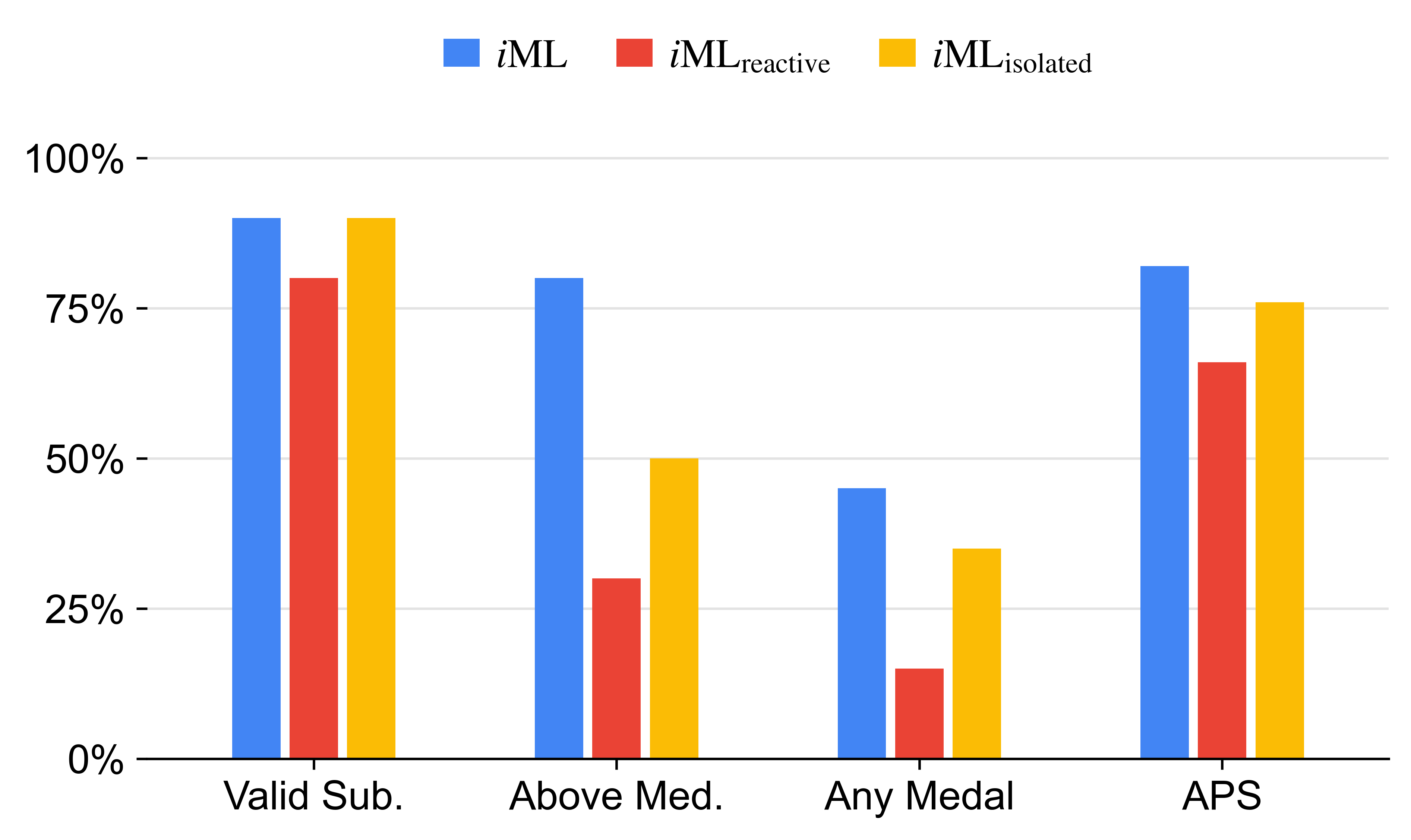}
    \caption{Impact of data profiling and knowledge retrieval on \tool's performance on \mlebench.}
    \label{fig:iml_variants2}
\end{figure}

As shown in Figure~\ref{fig:iml_variants2}, removing data profiling from the \textit{Guideline Agent} in \tool$_{\text{reactive}}$ led to a substantial performance degradation, with APS dropping from 0.82 to 0.66. Although this variant still maintained an 80\% valid submission rate, its competitive quality deteriorated sharply, as reflected by the decline in \textit{Above Median} submissions from 80\% to 30\%. This result suggests that, without empirical profiling, the coding agents struggle to synthesize code under sufficiently explicit architectural constraints and must instead rely on weaker inferences from the raw task description alone.

More specifically, the absence of profiling prevented the planning process of crucial information about the observed properties of the dataset, such as schema structure, feature types, and data quality patterns. As a result, the generated guidance becomes less effective at aligning preprocessing decisions with downstream modeling requirements. In turn, \tool tends to fall back on more generic implementation strategies that remain executable, but are less well matched to the actual data distribution and therefore less competitive.
These results indicate that data grounding is not merely helpful for stabilizing execution, but is a major driver of high-tier competitive quality.

\textbf{Impact of Knowledge Retrieval.}
To evaluate the contribution of retrieved task-relevant knowledge, we evaluated \tool$_{\text{isolated}}$ (\textit{No knowledge retrieval}), a degraded variant in which the Knowledge Retriever is removed. In this setting, the system relies only on the task description and empirical data profiling without external retrieval of preprocessing strategies or domain constraints.

As shown in Figure~\ref{fig:iml_variants2}, removing knowledge retrieval led to a noticeable decline in competitive performance while preserving executability. In particular, \tool$_{\text{isolated}}$ maintained the same 90\% valid submission rate as the full system, but its \textit{Above Median} rate dropped by 37.5\%, its \textit{Any Medal} rate decreased from 45\% to 35\%, and APS slightly declined  by about 8\%, relatively.

These results suggest that empirical profiling alone is insufficient for constructing highly competitive pipelines. While the system can still generate executable solutions grounded in the observed properties of the dataset, the absence of retrieved task-relevant knowledge weakens its ability to identify preprocessing strategies, architectural choices, and implementation patterns that are better aligned with the broader task context. Consequently, the generated pipelines remain functional but become less competitive compared to those produced by the full framework.

More broadly, the results indicate that problem grounding in \tool emerges from the combination of empirical evidence and external task-relevant knowledge. Data profiling helps the system understand the observed structure of the dataset, while knowledge retrieval provides additional guidance about how that structure should be processed and modeled in practice.

\subsubsection{Exploration-oriented Analysis}

Finally, we analyze how \tool supports broad and up-to-date exploration of valid solutions, rather than committing early to a single narrow modeling path. This property is particularly important because \tool is designed to explore multiple implementation tracks and incorporate external knowledge when appropriate.

\textbf{Impact of Multi-track Exploration.}
To verify the impact of \tool's dynamic selection capability, we conducted a comparative analysis between \tool and three ``single-implementation-track'' variants, each restricted to a fixed modeling paradigm.

\begin{figure}
    \centering
    \includegraphics[width=1.0\linewidth]{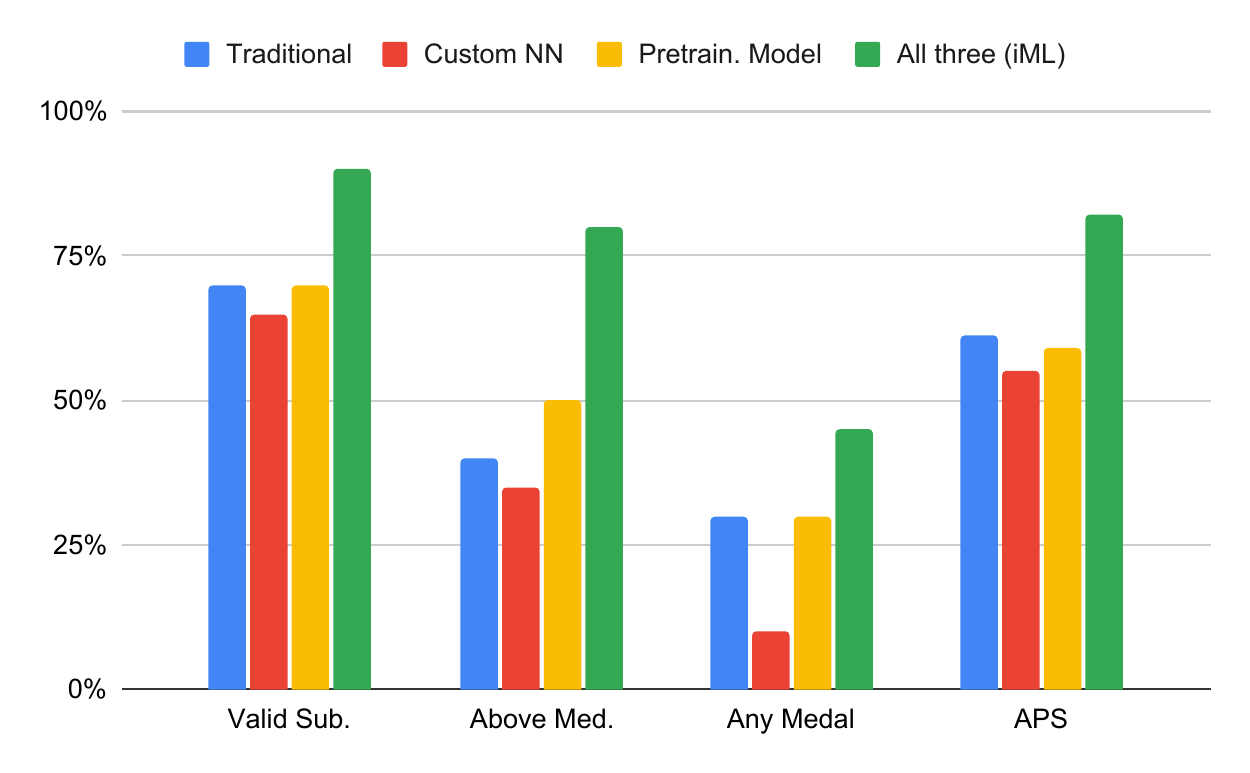}
    \caption{Performance of \tool's variants with different model selection strategies on \mlebench.}
    \label{fig:model_selection}
\end{figure}

Our results in Figure~\ref{fig:model_selection} show that restrictive configurations suffer from inherent inductive bias limitations. The \textit{Traditional} track, exploring only traditional ML approaches, while robust for standard tabular datasets, achieved an APS of 0.61 and an \textit{Any Medal} rate of only 30\%, as it lacks the specialized flexibility required for complex image or text tasks. Similarly, the \textit{Custom NN} variant recorded the lowest APS (0.55), showing the high failure rate (65\% valid submission) and resource intensity associated with architecting neural networks from scratch without the fallback of established baselines. While \textit{Pretrained Model} showed promise with a 50\% \textit{Above Median} rate, its rigid dependency on existing checkpoints prevented it from optimizing for niche distributions where traditional algorithms might excel.

The full version of \tool significantly outperformed the best single-track variant, achieving an APS of 0.82 and an \textit{Any Medal} rate of 45\%. This improvement is a consequence of \tool's ability to perform cross-track reasoning: by exploring \textit{all} three strategies and employing Evaluation Agent for final selection, \tool ensures that the chosen pipeline is empirically optimal for the task's specific meta-features.

This dynamic selection effectively eliminates the performance collapse seen in fixed-strategy baselines. Compared to letting an LLM decide a single approach upfront, this strategy reduces mode-collapse and early commitment risk: each track is executed and evaluated empirically under the same data/meta-feature, so the final choice is driven by measured performance rather than prompt-sensitive reasoning or prior assumptions about which track ``should'' work.

\textbf{Impact of Search Engine / Retrieval Backend.}
To evaluate the contribution of external retrieval to \tool's exploration capability, we compared the full \tool framework against \tool$_{\text{retrieval-free}}$, a degraded variant in which the search engine is removed. 

\begin{figure}
    \centering
    \includegraphics[width=1.0\linewidth]{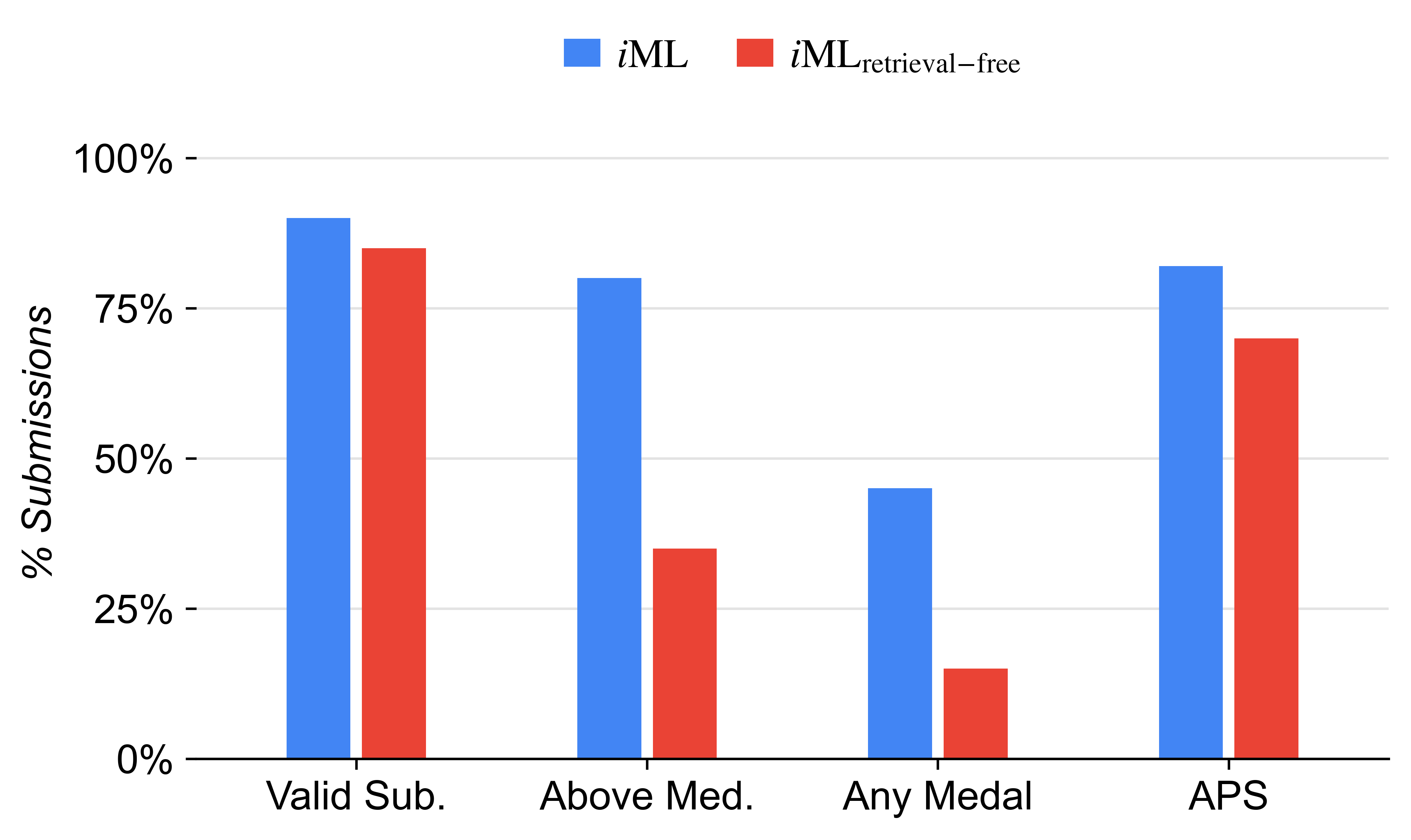}
    \caption{Impact of removing search engine on \tool's performance on \mlebench.}
    \label{fig:retrieval-free}
\end{figure}

Figure~\ref{fig:retrieval-free} shows that disabling retrieval leads to a consistent performance drop across all major metrics. In particular, \tool$_{\text{retrieval-free}}$ reduced the \textit{Valid Submission} rate from 90\% to 85\%, the \textit{Above Median} rate from 80\% to 35\%, the \textit{Any Medal} rate from 45\% to 15\%, and APS from 0.82 to 0.70.

These results show that external retrieval contributes more than a marginal improvement in executability. While the \tool$_{\text{retrieval-free}}$ preserves a quite high valid submission rate, its competitive quality declines sharply. This suggests that the search engine plays a critical role in helping \tool discover stronger and more up-to-date solution candidates, including relevant architectures, pretrained checkpoints, and implementation patterns that are difficult to recover from the backbone LLM's internal prior knowledge alone.

\textbf{Impact of Aggregation Logic.}
To evaluate the effectiveness of aggregation strategies, we compared the default \textit{Best Valid} selection strategy, which identifies the single highest-performing track among the three model implementation tracks during internal validation, against three common ensemble techniques: \textit{Simple Average}, \textit{Stacking}, and \textit{Boosting}.

\begin{figure}
    \centering
    \includegraphics[width=1.0\linewidth]{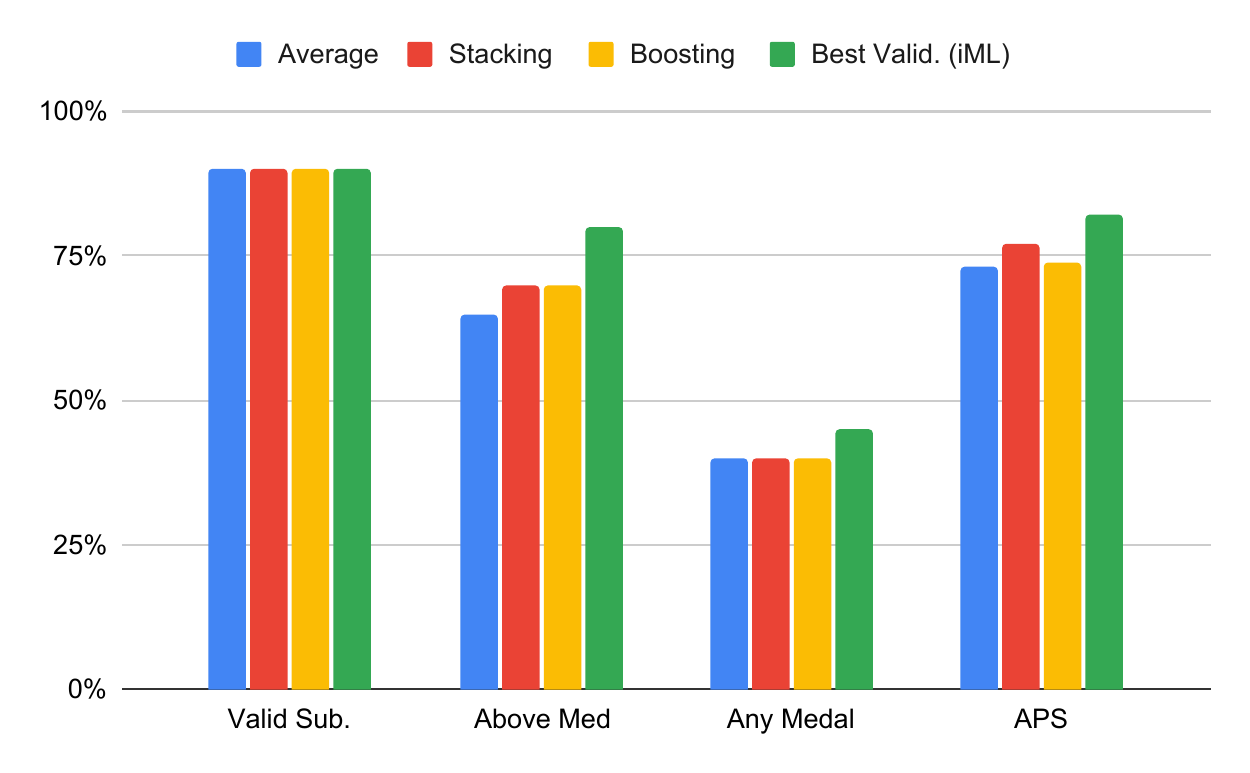}
    \caption{Performance of \tool's variants with representative ensemble techniques on \mlebench.}
    \label{fig:ensemble}
\end{figure}

In Figure~\ref{fig:ensemble}, while all ensembling variants maintained a consistent 90\% valid submission rate, they notably underperform relative to the single-best selection strategy in high-tier competitive metrics. Specifically, \tool with \textit{Best Valid} configuration achieved an APS of 0.82 and an \textit{Any Medal} rate of 45\%, whereas the strongest ensembling variant, \textit{Stacking}, reached an APS of only 0.77 and a 40\% medal rate.

This performance gap suggests that in complex ML tasks, the empirical selection of a single, highly optimized pipeline anchored by the strategic blueprint is more effective than naive multi-track aggregation. Although modular ensembling achieves high robustness with \textit{Above Median} rates of 65\% to 70\%, it appears to weaken the task-specific optimizations of the global optimum, leading to a performance decrease. Consequently, while \tool's architecture supports ensembling, we recommend prioritizing refined track selection over aggregation for competitive tasks where depth of optimization outweighs simple architectural diversity.

\textbf{Impact of Backbone LLM.}
To evaluate the impact of the underlying LLM on \tool's performance, we evaluated \tool with different backbone LLMs ranging from the specialized \textit{Qwen 2.5 Coder 7b} to frontier-class models like \textit{GPT-5 mini} and \textit{Gemini 2.5}.

\begin{figure}
    \centering
    \includegraphics[width=1.0\linewidth]{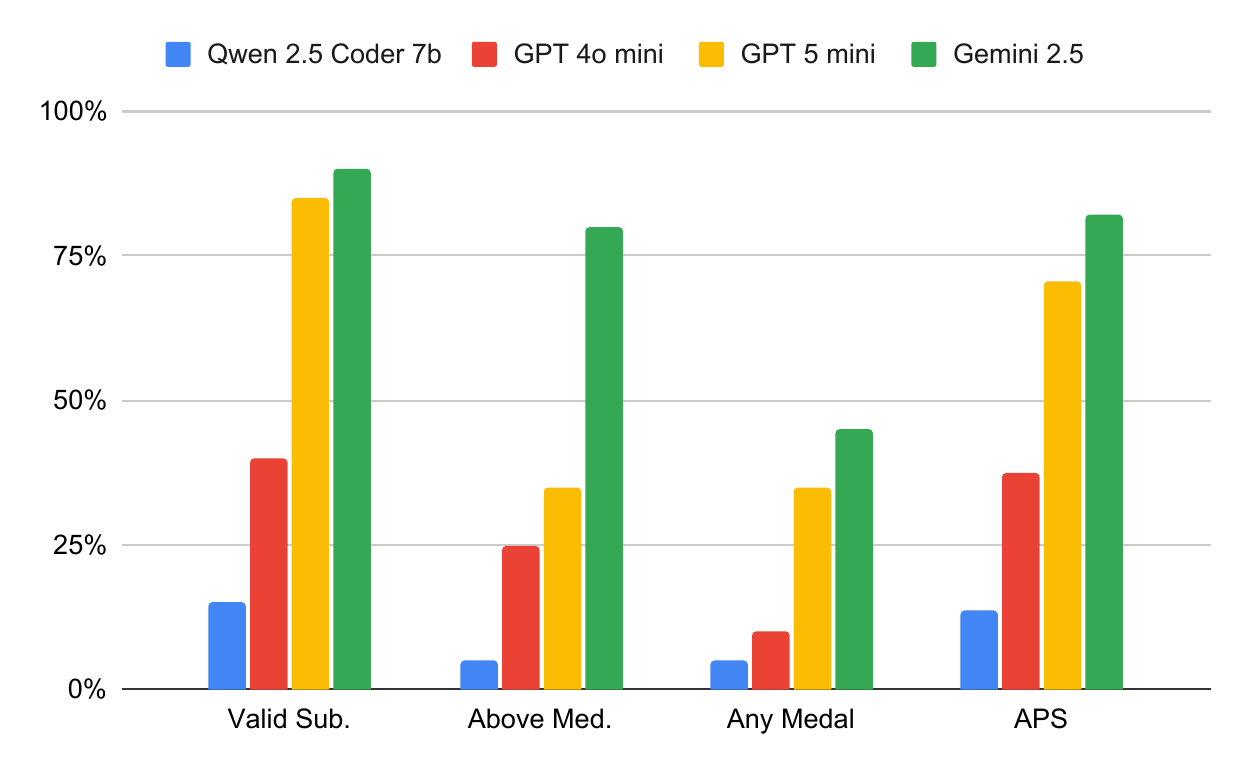}
    \caption{Performance of \tool's variants with different backbone LLMs on \mlebench.}
    \label{fig:backbone_model}
\end{figure}

In Figure~\ref{fig:backbone_model}, Gemini 2.5 serves as the optimal backbone for \tool, achieving the highest APS of 0.82 and a dominant 80\% \textit{Above Median} rate. This suggests that high-level reasoning capabilities are critical for \tool's coding and planning agents to synthesize the complex architectural invariants and correct code required for competitive medal achievement.

With \textit{Qwen 2.5 Coder 7b}, despite its coding specialization, \tool achieved only a 15\% valid submission rate and a negligible APS of 0.14, indicating that code generation in isolation is insufficient for navigating end-to-end ML pipelines without stronger strategic reasoning. Furthermore, while \textit{GPT-5 mini} matched Gemini's 85\% valid submission rate, it suffered a significant drop in competitive quality, securing only a 35\% \textit{Above Median} rate compared to Gemini's 80\%. This indicates that while intermediate frontier models can follow the executable integration process to produce runnable scripts, they often fail to implement the deep, task-specific optimizations necessary for top-tier performance.

These results suggest that broad exploration in \tool depends not only on access to multiple candidate solution paths, but also on the backbone model's ability to reason over, refine, and select among them effectively.

\textbf{\textit{Overall}}, \tool's performance arises from the joint contribution of executability-oriented mechanisms, problem-grounded planning, and broad solution exploration. The executability-oriented analyses show that dynamic verification, modular integration, and self-correction are essential for producing feasible pipelines. The problem-grounding analysis shows that empirical profiling and knowledge retrieval are major drivers of competitive quality. The exploration-oriented analyses show that \tool benefits substantially from searching across multiple modeling tracks, while also depending on strong retrieval and reasoning components to exploit that broader search space effectively.

%% file: 6.results/RQ3.tex
\subsection{RQ3. Sensitivity Analysis}
\subsubsection{Task Description Granularity}

\begin{figure}
    \centering
    \includegraphics[width=1.0\linewidth]{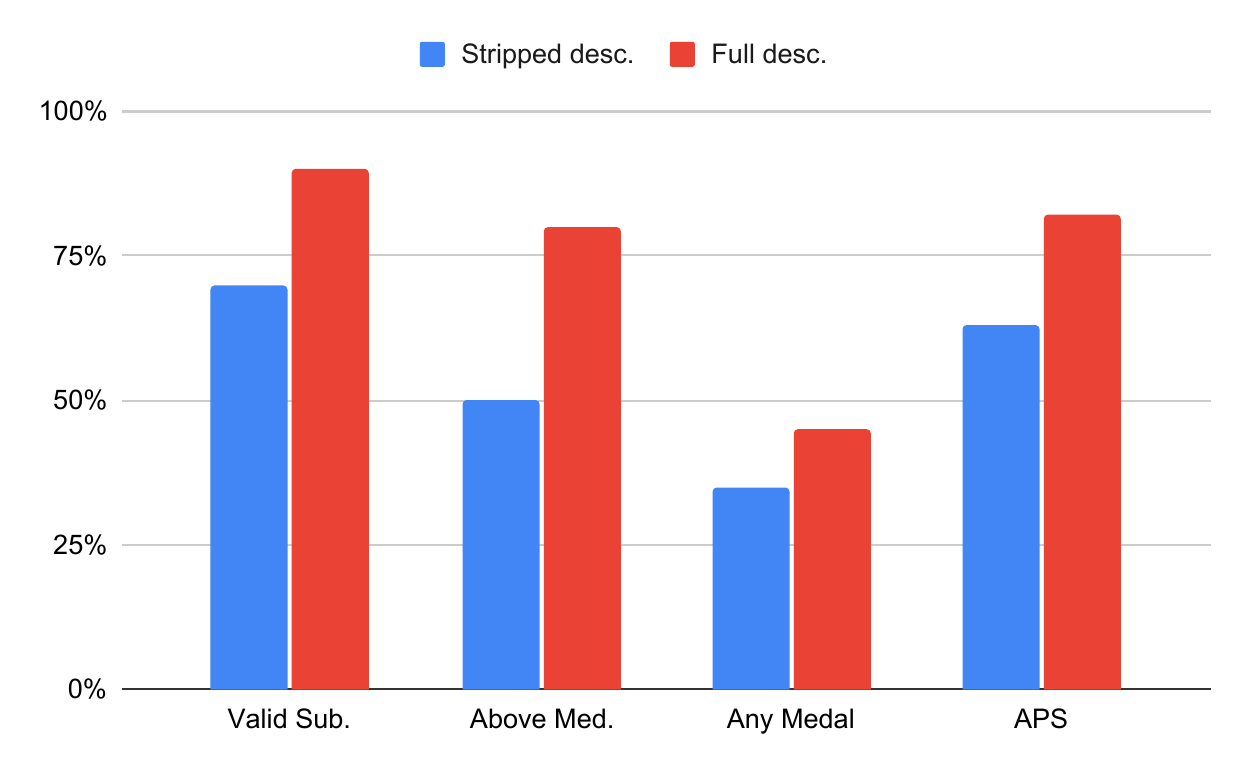}
    \caption{Performance of \tool when providing full and stripped task descriptions on MLE Bench}
    \label{fig:task_desc}
\end{figure}

% https://docs.google.com/spreadsheets/d/11rWpSZJdHk56ulwIZYiWF95gP7_pRkcl58DscqdDDQc/edit?gid=344280593#gid=344280593

To evaluate the robustness of \tool against ambiguous user intent, we conducted an experiment comparing its performance when provided with comprehensive competition descriptions versus minimalist, stripped descriptions. To ensure the objectiveness and realistic nature of this evaluation, we standardized the extraction of stripped descriptions by systematically removing all domain-specific context, feature explanations, and instructional guidance, replacing them with a singular, task-agnostic prompt: \textit{``Analyze the provided data files and execute the appropriate predictive task for the target variable''.} This methodology simulates a ``zero-knowledge'' scenario that forces the framework to move beyond textual reasoning and rely primarily on autonomous empirical profiling of the raw data files to recover the underlying problem structure and target semantics.

As shown in Figure~\ref{fig:task_desc}, \tool demonstrates remarkable robustness to reduced task descriptions. While with full description, \tool achieves an APS of 0.82 and a valid submission rate of 90\%, with stripped description, our approach still maintains a highly competitive APS of 0.63 and a 70\% valid submission rate. The relatively mild degradation in performance suggests that \tool does not rely exclusively on textual heuristics. Instead, \textit{Guideline Agent} effectively utilizes empirical meta-features ($F_{meta}$), such as data types, class distributions, and correlation matrices, to infer the underlying problem semantics (e.g., classification vs. regression) and select appropriate optimization metrics.

Furthermore, even with stripped task description, \tool still achieved \textit{Above Median} performance in 50\% of the tasks and secured medals in 35\% of cases. This stability is primarily attributed to \textit{Code-Guided Strategic Planning} initialization phase, where \tool executes profiling scripts to anchor its strategic planning in the physical reality of the dataset rather than probabilistic textual inference. These results indicate that \tool is not merely a prompt-dependent wrapper but a robust engineering system capable of navigating under-specified tasks by extracting strategic insights directly from the data artifacts

% The full description outperformed the stripped description across all metrics: Above Med. increased from 50\% to 80\% (+30 percentage points), Valid Sub. increased from 70\% to 85\% (+15\%), and APS increased from 0.63 to 0.77. Notably, even with the description stripped, iML still achieved a Valid Sub. of 70\%, Above Med. of 50\%, and an APS of 0.63, demonstrating that the Profile Agent and Guideline Agent compensated for the missing information in the original description.

% When the description was stripped, the Profile Agent analyzed the actual data (schema, distribution, missing values, data types) to add information missing from the description.

% The Guideline Agent synthesized the profile and description to create a detailed guideline, helping the coder agents understand the task even with an incomplete description.

% As a result, even with a stripped description, iML still achieved a Valid Sub. of 70\% and Above Med. of 50\%—significantly higher than relying solely on a stripped description.

\subsubsection{Machine Learning Task Type}

\begin{figure}
    \centering
    \includegraphics[width=1.0\linewidth]{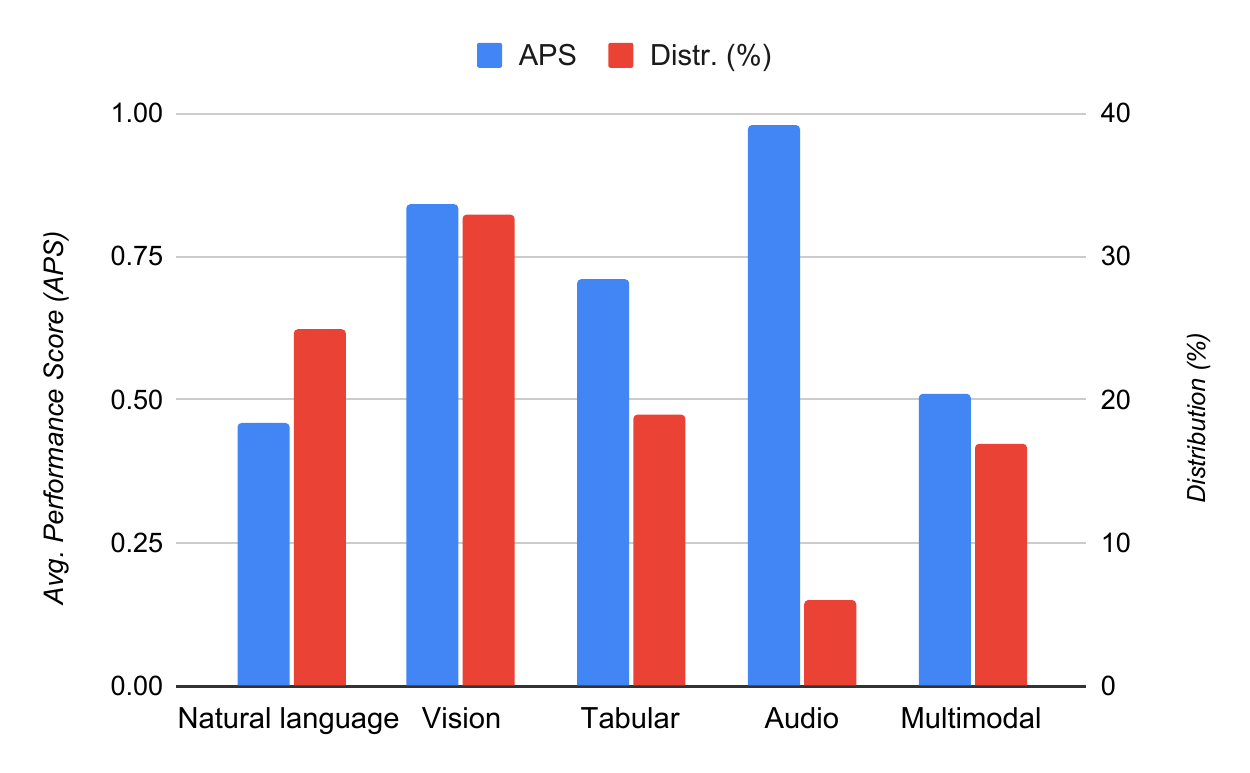}
    \caption{Performance of \tool by task types on \mlebench}
    \label{fig:ml_type}
\end{figure}

In this experiment, we investigated \tool's performance by five distinct data modalities. Figure~\ref{fig:ml_type} shows that \tool maintains high effectiveness across a broad spectrum of data modalities, confirming the robustness of its \textit{Code-Guided Planning} phase. As shown in Figure~\ref{fig:ml_type}, \tool achieved its highest APS in Audio tasks (0.98), despite these tasks representing a specialized subset (3\%) of the total distribution. This near-perfect score suggests that when \textit{Guideline Agent} identifies structured signal data, it effectively utilizes custom neural networks or specialized pretrained models to architect high-precision extractors.

\tool also exhibited strong performance in Vision (0.83 APS) and Tabular (0.72 APS) modalities, which collectively constitute nearly half of the experimental workload. The high score in Vision tasks specifically highlights the success of \tool's capability to surpass traditional ``black-box'' tabular tools by synthesizing native neural architecture. Conversely, the Multimodal category, which requires the complex integration of disparate data sources, presented the greatest challenge, yielding an APS of only 0.51. 
This degradation show that while \tool's coding agents can successfully bind variables from multiple modules, the strategic overhead of orchestrating cross-modal feature engineering remains a frontier for further optimization.

Overall, the consistency of the APS across these varied modalities underscores that \tool's architecture allows it to adapt its inductive biases dynamically, ensuring it remains competitive even when transitioning between different modalities like natural language processing and traditional statistical modeling.

% https://docs.google.com/spreadsheets/d/11rWpSZJdHk56ulwIZYiWF95gP7_pRkcl58DscqdDDQc/edit?gid=635298296#gid=635298296

%% file: 6.results/RQ4.tex
\subsection{RQ4. Efficiency Analysis}

In this work, all experiments were conducted on a server running with NVIDIA P100 GPU. We employed Gemini 2.5 Flash as the default back-bone LLM in \tool.
On average, \tool took about 3.5 hours, 185K tokens, and only 0.20 US dollar to produce a ready-to-deploy ML model.
During produce a model, \tool spent the largest proportion (85\%) of the total running time to execute, verify, and debug the pipeline $\mathcal{S}^*$ (Phase III: Code-Verifiable Integration). Meanwhile, \tool took only 6 minutes to generate a blueprint, and 20 minutes to generate the first version of the pipeline's script.

% https://docs.google.com/spreadsheets/d/11rWpSZJdHk56ulwIZYiWF95gP7_pRkcl58DscqdDDQc/edit?gid=1135961461#gid=1135961461

%% file: 6.results/threats.tex
\subsection{Threats to Validity}

The main threats to the validity of our work are consisted of internal, external, and construct validity threats.

\textbf{Threats to Internal Validity}.
One potential threat concerns the correctness of the implementation across our multi-agent framework. To mitigate this risk, we modularized the source code by separating Guideline, Coding, and Assembler, and the other agents, and make the framework publicly available to allow researchers to reproduce and verify our results~\cite{website}. 
Another internal threat relates to the stochasticity of LLMs, which can influence the consistency of code generation and debugging. To reduce this threat, we utilized consistent temperature settings and systematically evaluated the impact of the self-correction budget $K$ across multiple iterations to identify the optimal efficiency frontier. 
Finally, the choice of hyperparameters, such as the maximum debugging rounds and prompt templates, may influence predictive performance. We addressed this by conducting sensitivity analyses across diverse task distributions to establish robust default configurations.

\textbf{Threats to External Validity}.
A potential external validity threat arises from our reliance on specific LLM backbones (e.g., Gemini-2.5-Flash) for strategic planning and code generation. Variations in the quality of these models may affect the generalizability of our architectural advantages. To mitigate this, we evaluated \tool and all LLM-based baselines using the same backbone model under identical resource constraints to ensure fair comparison. We further conducted a cross-backbone comparison including GPT-5 mini and Qwen 2.5 Coder to demonstrate framework resilience.
Another threat concerns the generalization of our findings to diverse dataset domains. We addressed this by evaluating \tool on two comprehensive benchmarks, \mlebench and \imlbench, spanning multiple modalities including tabular, vision, natural language, audio, and multimodal tasks.

\textbf{Threats to Construct Validity}.
A potential threat to construct validity lies in the choice of evaluation metrics, such as medal-based rankings and normalized performance score. To mitigate this, we adopted standard competition metrics (e.g., Quadratic Weighted Kappa for \textit{PetFinder} task) and verified submission validity through the official grading systems of the benchmarks. 
Additionally, our use of \textit{Task Description Granularity} as a proxy for robustness to ambiguous intent may not capture all complexities of human-agent interaction. However, by quantifying the performance between full and stripped descriptions, we provide a concrete measure of the system's ability to rely on empirical data profiling over textual prompts. 

%% file: 8.conclusion.tex
\section{Conclusion}

In this work, we introduce \tool, a novel code-driven multi-agent framework that redefines AutoML through a code-guided, modular, and verifiable paradigm. By decoupling high-level strategic reasoning from low-level code implementation, \tool addresses the critical challenges of hallucinated logic and logic entanglement that frequently destabilize current monolithic AutoML agents. Our evaluation on the \mlebench and the newly introduced \imlbench shows that \tool significantly outperforms SOTA baselines, achieving a valid submission rate of 90\% and a competitive medal rate of 45\% on complex, real-world Kaggle competitions.

The results highlight the synergy of \tool's three core pillars: \textit{Code-Guided Planning}, \textit{Code-Modular Implementation}, and \textit{Code-Verifiable Integration}. Our results confirm that the \textit{Guideline Agent} provides essential strategic depth through autonomous empirical profiling, without which competitive performance drops significantly. Meanwhile, the \textit{Debugging Agent} ensures operational resilience by enforcing strict interface contracts and successfully repairing 45\% of initially failing scripts via dynamic debugging. Furthermore, \tool shows remarkable semantic autonomy, maintaining robust performance even when provided with stripped task descriptions, by grounding its planning in physical data reality.

Despite these advancements, the degradation in performance on multimodal tasks suggests a frontier for future research in cross-modal feature orchestration. Nonetheless, \tool represents a transformative step toward reliable, human-competitive autonomous engineering, establishing a verifiable foundation for the next generation of ML systems.